\documentclass[10pt,twocolumn,letterpaper]{article}

\usepackage{cvpr}              
\definecolor{cvprblue}{rgb}{0.21,0.49,0.74}
\usepackage[pagebackref,breaklinks,colorlinks,allcolors=cvprblue]{hyperref}

\usepackage{amsthm,mathtools,bm}
\usepackage{multirow,array,tabularx}
\usepackage{float,algpseudocode}
\usepackage{microtype}

\floatstyle{ruled}
\newfloat{algorithm}{tbp}{loa}
\floatname{algorithm}{Algorithm}
\newtheorem{proposition}{Proposition}
\setlist[itemize]{leftmargin=*,nosep}
\setlist[enumerate]{leftmargin=*,nosep}
\graphicspath{{figures/}}

\def\confName{CVPR}
\def\confYear{2027}

\title{RevealBelief: Calibrated Spatial Reasoning from Revealing Views}

\author{Xuanchen Wang \enskip Heng Wang  \enskip Weidong Cai \\
University of Sydney \\
{\tt\small xwan0579@uni.sydney.edu.au}\\
{\tt\small \{heng.wang, tom.cai\}@sydney.edu.au}
}

\begin{document}
\maketitle

\begin{abstract}
Spatial reasoning from a few images requires probabilities for relations that the images do not
resolve. Imagined viewpoints help inspect a proposed scene, while an acquired viewpoint provides
evidence about which scene is present. We study calibrated spatial reasoning under partial
observation: a system must answer related questions from one belief and revise those answers when
new evidence arrives. We propose RevealBelief, which distills a teacher with additional revealing
views into a question-independent scene distribution. A joint answer loss couples related events.
Shared scene samples support geometric queries and a compact rendered-language readout. On ScanNet,
RevealBelief lowers Brier score by $6.8\%$ and selective risk by $16.8\%$ against scene-likelihood
training; its Brier advantage increases to $13.9\%$ after one acquired view. Across five held-out
query families it improves accuracy over answer distillation by $4.6$ points. End-to-end language
evaluation, doubled-update controls, and exact-posterior diagnostics connect these gains to belief
learning, scene reuse, and conditional visual computation.
\end{abstract}

\section{Introduction}
\label{sec:intro}
\begin{figure*}[t]
\centering
\includegraphics[width=\textwidth]{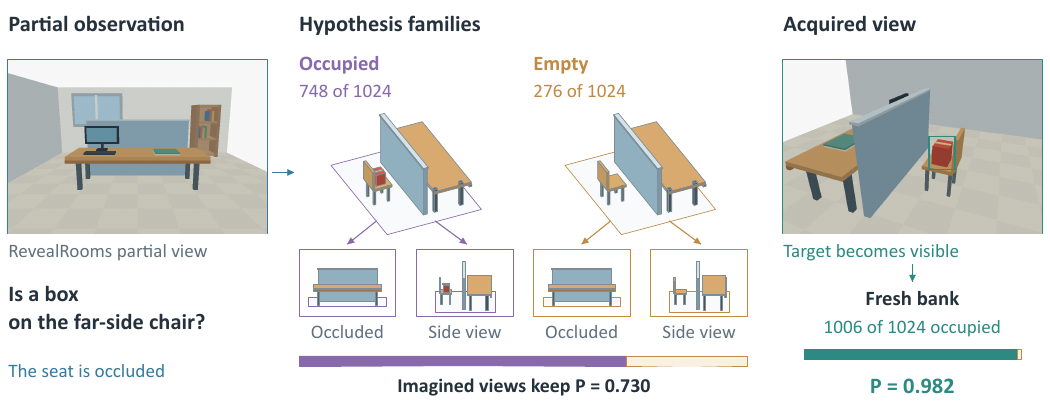}
\caption{\textbf{Inspect hypotheses; update from observations.} A RevealRooms episode leaves the
  target region occluded. The probability bank aggregates the hypotheses satisfying the query; their
  rendered views expose the geometry without adding observations. A newly acquired viewpoint
  distinguishes the alternatives and refreshes the bank. Counts refer to the $1{,}024$ probability
  samples; the $32$ rendered samples are a nested subset. All panels share the same scene and camera
  frame.}
\label{fig:teaser}
\end{figure*}

A chair behind a desk may hold an object that the current camera cannot see. Moving an imagined
camera around one guessed room makes that guess easy to inspect; it does not establish whether the
object is there. Several scenes may explain the same observed pixels. \Cref{fig:teaser} shows the
corresponding inference problem: predict an event from the probability of its compatible scenes, use
imagined views to reason within those scenes, and update their distribution after an acquired view.

World-model reasoners generate viewpoints for spatial inference~\cite{mindjourney,avic,astra},
invoke visual tools~\cite{act2see}, or navigate reconstructed environments~\cite{v2w}. Multi-view
generators increasingly support these computations~\cite{cat3d,stablevc}. Recent probing shows that
lower answer entropy after imagination can coexist with unreliable
verification~\cite{probingworldmodels}. The missing quantity is the probability of the underlying
scene: agreement among views rendered from one proposal is conditional agreement, whereas an answer
about the environment must average over competing proposals.

A completion model can omit a plausible arrangement, misallocate probability, or select different
implicit scenes for related questions. Calibration does not supply missing geometry, and answer
heads must extrapolate to new operators. We instead train a normalized distribution over scene
codes. A teacher observes additional acquired views during training; its distribution supervises a
student that receives the partial observation. A joint answer-signature objective emphasizes the
probability of related spatial events without discarding the scene structure.

Both readouts reuse the same sampled scenes: geometric execution averages their events, while a
compact subset supplies rendered views to a language solver. Additional measurements refresh the
belief.

The clearest gain appears when new evidence arrives. On ScanNet, the Brier reduction over
Scene NLL grows from $6.8\%$ before acquisition to $13.9\%$ afterward. Across five
leave-one-family-out settings, the same representation improves accuracy by $4.6$ points over answer
distillation and $2.2$ over scene likelihood. These comparisons include random-view-subset and
doubled-update controls. Video transfer and end-to-end composed-language results connect the
calibrated geometric protocol to visual reasoning. Benchmark specifications and scoring routines accompany the supplement. Our contributions are:
\begin{itemize}
\item We propose RevealBelief, which learns one question-independent scene distribution through
revealing-view distillation and joint answer supervision, and supports geometric and visual readouts
through shared hypotheses.
\item We introduce RevealRooms with natural procedural episodes and observation-equivalent families
with known posteriors, alongside a grouped ScanNet protocol for evaluating outcome calibration.
\item We demonstrate gains in spatial reasoning with shared-architecture controls, acquired-view
updates, five-family reuse, and end-to-end language evaluation; exact integration separates belief
error from sampling error.
\end{itemize}

\section{Related Work}
\label{sec:related}
\paragraph{Spatial Reasoning and Evaluation.}
SpatialVLM~\cite{chen2024spatialvlm}, SpatialRGPT~\cite{cheng2024spatialrgpt}, and SAT~\cite{sat}
develop spatial supervision; ScanQA~\cite{scanqa}, SQA3D~\cite{sqa3d}, and OpenEQA~\cite{openeqa}
connect questions to embodied environments. VSI-Bench~\cite{vsibench}, MMSI-Bench~\cite{mmsibench},
and MindCube~\cite{mindcube} extend evaluation to video and mental scene models. We study event
probabilities under incomplete evidence and reuse across questions.

\paragraph{Scene Representations and Visual Computation.}
SceneScript~\cite{avetisyan2024scenescript}, CUT3R~\cite{wang2025cut3r}, and VGGT~\cite{vggt}
provide structured or persistent geometry. Scene completion~\cite{sscnet}, object-centric
learning~\cite{slotattention}, and query-based generation~\cite{gqn} motivate explicit hypotheses;
DFM~\cite{dfm}, CAT3D~\cite{cat3d}, and Stable Virtual Camera~\cite{stablevc} improve novel views.
MindJourney~\cite{mindjourney}, AVIC~\cite{avic}, Astra~\cite{astra}, Act2See~\cite{act2see},
V2W~\cite{v2w}, and ImagiRep~\cite{imagirep} use visual representations as computation. Our two
readouts separate that computation from the scene probability it conditions upon.

\paragraph{Beliefs and Observation.}
3D-Belief~\cite{belief3d} represents plausible completions and updates, while Language as a
Sensor~\cite{languagesensor} and PUF~\cite{yang2026puf} fuse spatial information.
LEDGERMIND~\cite{ledgermind} tracks provenance; covariance
intersection~\cite{julier1997covintersection} addresses dependence. Bayes' Rays~\cite{bayesrays} and
FisherRF~\cite{fisherrf} study reconstruction uncertainty. KnowNo~\cite{knowno}, Explore until
Confident~\cite{exploreconfident}, and ACT3D~\cite{act3d} connect uncertainty or exploration to
further observations. We learn reusable scene mass under fixed acquisition schedules.

\paragraph{Calibration and Privileged Supervision.}
SpatialUncertain~\cite{spatialuncertain}, VirtueBench~\cite{virtuebench}, reasoning-overconfidence
studies~\cite{costreasoning}, evidence-order calibration~\cite{evidenceorder}, and VPAC~\cite{vpac}
motivate evidence-sensitive decisions. We compare self-consistency~\cite{wang2023selfconsistency}
using proper scores~\cite{brier1950,gneiting2007proper},
reliability~\cite{dawid1982calibration,murphy1973decomposition,guo2017calibration}, and selective
risk~\cite{selectivenet}. Ensembles~\cite{lakshminarayanan2017ensembles}, uncertainty
taxonomies~\cite{kendall2017uncertainties}, and shift studies~\cite{ovadia2019datasetshift} address
complementary uncertainty sources.
Distillation~\cite{hinton2015distillation,lopezpaz2016generalized,learningbycheating}, particle
methods~\cite{gordon1993particle,doucet2001smc}, and multivariate
scores~\cite{scheuerer2015variogram} supply the foundations for our learned scene interface and
finite-bank analysis.

\section{RevealBelief}
\label{sec:method}
\begin{figure*}[t]
\centering
\includegraphics[width=\textwidth]{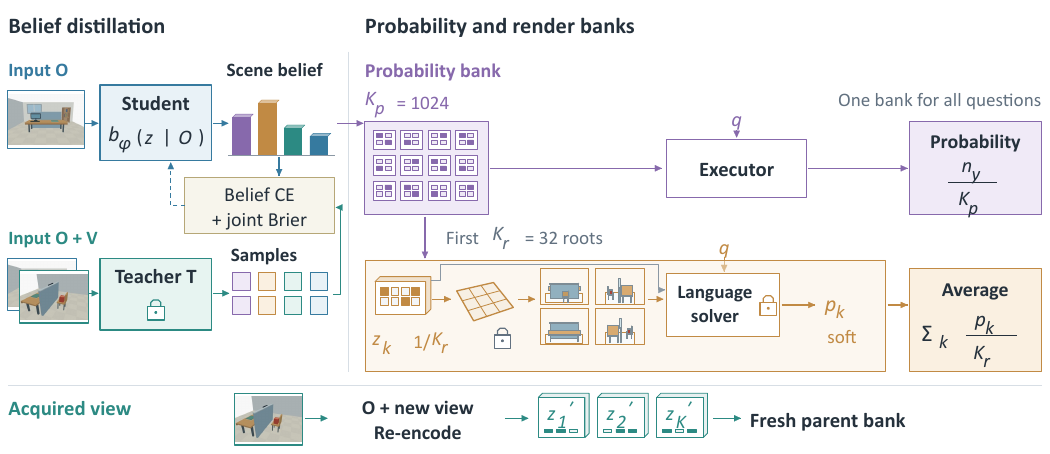}
\caption{\textbf{One learned belief, two readouts.} A revealing-view teacher and a joint answer loss
  supervise the partial-view student. Its $K_p=1{,}024$ samples support exact template execution;
  the first $K_r=32$ samples additionally supply imagined views and soft language predictions. Each
  path averages once; $n_y$ counts executor answers equal to $y$. Acquired images refresh both
  banks. Locks indicate frozen modules.}
\label{fig:framework}
\end{figure*}

RevealBelief distills a multi-view teacher into a reusable partial-view scene belief
(\cref{fig:framework}). Related questions supervise its joint event probabilities; geometric and
language readouts average once over scene hypotheses.

\subsection{Problem Setup}
An acquired set $\mathcal{O}=\{(x_i,c_i)\}$ contains RGB images and camera parameters in a common
episode frame. A hypothesis $z$ records object presence, semantic categories, quantized geometry,
surface-shape codes, and appearance. Its 64 potential slots include empty slots, so the active
inventory is inferred. A frozen shape decoder converts categorical surface codes into renderable
geometry. The same state represents procedural rooms and annotated real scans.

A set encoder and autoregressive decoder define the question-independent belief:
\begin{equation}
 b_\phi(z\mid\mathcal{O})=
 \prod_{\ell=1}^{L}b_{\phi,\ell}(z_\ell\mid z_{<\ell},\mathcal{O}).
 \label{eq:belief}
\end{equation}
Each conditional is a normalized categorical distribution. Canonical serialization gives each state
one code sequence. The representation supports geometric execution of the declared query family;
composed language queries use a learned solver. We evaluate answer-neutral questions whose selection
does not disclose the hidden scene. Details of the state, geometry tolerances, and solver interface
appear in Appendix~B.

\subsection{Revealing-View Distillation}
The teacher $T_\psi(z\mid\mathcal{O},\mathcal V)$ receives $\mathcal{O}$ and four additional
acquired training views $\mathcal V$ (\cref{fig:framework}, left). We train it by scene-state
negative log likelihood using complete procedural labels or reference-supported scan annotations,
then freeze its parameters. Random view subsets expose the teacher to both identifying and ambiguous
inputs. Its normalized distribution supplies soft targets rather than substituting the realized
scene for its predictive uncertainty.

The student minimizes the cross-entropy:
\begin{equation}
 \mathcal L_{\rm belief}=
 \mathbb{E}_{\mathcal{O},\mathcal V}
 \left[-\sum_zT_\psi(z\mid\mathcal{O},\mathcal V)
 \log b_\phi(z\mid\mathcal{O})\right].
 \label{eq:distill}
\end{equation}
Eight teacher samples estimate this expectation. Scene NLL shares the vocabulary, labels, partial
inputs, student architecture, and update budget, isolating the target.

Both networks use frozen DINOv2 ViT-L/14 features~\cite{dinov2}, a camera-aware set adapter, and a
12-layer scene decoder. Their common categorical output space includes uncertain presence and shape;
architecture and optimization details are in Appendix~B.

\begin{proposition}[Posterior Distillation]
\label{prop:distill}
For a finite scene space and one fixed data-generating law, suppose
$T_\psi(z\mid\mathcal{O},\mathcal V)=P(z\mid\mathcal{O},\mathcal V)$. Over unrestricted conditional
distributions, every population minimizer of \cref{eq:distill} equals $P(z\mid\mathcal{O})$ almost
surely.
\end{proposition}
The tower property averages the revealing-view posterior back to the partial-view posterior.
Appendix~A gives the proof, an imperfect-teacher bound, and the conditional-gradient
variance identity. Under the exact-teacher assumption, distillation and Scene NLL share
the same expected gradient at fixed student parameters. Their
stochastic gradients can nevertheless differ: averaging independent
teacher samples reduces the variance due to scenes that remain
ambiguous after the revealing views. With eight samples, this
conditional variance component is divided by eight, while variation
across observation--reveal pairs remains. Fully identifying reveals
recover the point-label gradient. The teacher-quality, data-scale,
and doubled-update controls evaluate the practical benefit of the
learned targets.


\subsection{Joint Answer Signatures}
A cluster $Q=(q_1,\ldots,q_m)$ asks related questions about shared geometry. Its answer signature is
$F_Q(z)=(f(z,q_1),\ldots,f(z,q_m))$, where $f$ executes the training predicates. The belief induces:
\begin{equation}
 p_\phi(\mathbf y\mid\mathcal{O},Q)=
 \sum_zb_\phi(z\mid\mathcal{O})\mathbf{1}[F_Q(z)=\mathbf y].
 \label{eq:joint}
\end{equation}
We optimize the expected joint Brier score:
\begin{equation}
 \mathcal L_{\rm joint}=
 \mathbb{E}_{\mathbf y^\star}
 \sum_{\mathbf y}
 \big(p_\phi(\mathbf y\mid\mathcal{O},Q)
      -\mathbf{1}[\mathbf y=\mathbf y^\star]\big)^2,
 \label{eq:jointbrier}
\end{equation}
using teacher-sampled signatures. The combined objective is
$\mathcal L_{\rm belief}+\lambda\mathcal L_{\rm joint}$.

Scene likelihood distinguishes full geometries; the joint score emphasizes their shared event
probabilities. For example, \emph{red left of blue} and \emph{blue right of red} collect the same
hypothesis mass, and an acquired observation revises them together. Joint scoring also distinguishes
beliefs whose marginal answers agree but co-occurrences differ. Appendix~B gives an
unbiased sampled estimator. We score joint probabilities separately from independently thresholded
decisions.

\subsection{Inference and Update}
Draw $K_p=1{,}024$ independent hypotheses from $b_\phi(\cdot\mid\mathcal{O})$. The first $K_r=32$
form a nested render subset. Both are reused across questions, with weight $1/K_p$ or $1/K_r$ in the
corresponding estimator. Independent draws that coincide retain their combined mass; extra
computational copies preserve their source weight.

\paragraph{Probability Estimation.}
A geometric predicate returns $f(z,q)\in\{0,1\}$. Writing $p=\mathbb{E}_b[f(Z,q)]$ and $\hat
p=K_p^{-1}\sum_k f(z_k,q)$ gives:
\begin{equation}
\mathbb{E}[(\hat p-y)^2\mid p,y]=(p-y)^2+\frac{p(1-p)}{K_p}.
\label{eq:bank}
\end{equation}
Thus sampling adds at most $1/(4K_p)$ to expected binary Brier, below $2.45\times10^{-4}$ at the
default budget. The unbiased diagnostic:
\begin{equation}
(\hat p-y)^2-\frac{\hat p(1-\hat p)}{K_p-1}
\label{eq:bank-correction}
\end{equation}
estimates the infinite-bank score; Appendix Table~A11 compares it with explicit large-bank
sampling. We retain raw pushforward probabilities for accuracy, Brier, calibration, and joint
consistency. Binary log-loss reporting uses $(K_p\hat p+1/2)/(K_p+1)$, with the same half-count
reporting kernel applied to soft-head predictions. Appendix~C.4 gives the
conventional-log-loss sensitivity and joint-signature rule.

\paragraph{Conditional Visual Computation.}
For a composed-language question, a frozen Qwen3-VL solver~\cite{qwen3vl} receives each
render-subset code and either no images or $M$ imagined views. The default $M=4$ uses four evenly
spaced azimuths and fixed camera settings (Appendix~B). A soft conditional
prediction $p_k(y\mid q)$ produces:
\begin{equation}
\widehat p(y\mid\mathcal{O},q)=\frac{1}{K_r}\sum_{k=1}^{K_r}p_k(y\mid q).
\label{eq:mixture}
\end{equation}
No within-hypothesis answer is thresholded before averaging. For each scalar answer probability in
this soft readout, the sampling term is $\operatorname{Var}_b[p_Z(y\mid q)]/K_r$. The visual budget
can change the solver's conditional answer, while scene weights remain fixed.

\paragraph{Acquired Evidence.}
A new acquired image extends $\mathcal{O}$, recomputes $b_\phi$, and refreshes the probability bank
and its render subset. Every related question then uses the updated distribution. Fixed acquisition
schedules isolate this update from camera-policy learning. Acquisition identifiers deduplicate
copied measurements; render caches are indexed by hypothesis and camera.

\paragraph{Real-Scene Instantiation.}
ScanNet RGB episodes use provided poses~\cite{scannet}. Training instance surfaces supply shape
codes through a frozen vector-quantized decoder~\cite{vqvae}; inference renders predicted codes. All
real-scene arms use Natural pretraining followed by grouped ScanNet fitting. The RGB-only video
protocol uses a common VGGT frontend~\cite{vggt}. Reference-supported regions define scored
outcomes, and a fixed episode frame keeps acquisitions comparable. Appendix~B
specifies geometry, camera alignment, and input metadata.

\subsection{Error Budget}
Let $p^\star$ denote the true answer distribution, $\mathcal S$ the represented hypothesis support,
$\eta=P(Z\notin\mathcal S\mid\mathcal{O})<1$, and $P_{\mathcal S}$ the true posterior conditioned on
$\mathcal S$. For predicted mass $\widehat\pi$, solver kernels $p_z$, and reference kernels $f_z$,
define $\epsilon_{\rm mass}=\operatorname{TV}(\widehat\pi,P_{\mathcal S})$ and $\epsilon_{\rm
solve}=\sum_{z\in\mathcal S}\widehat\pi(z)\operatorname{TV}(p_z,f_z)$. Then:
\begin{equation}
 \operatorname{TV}(\widehat p,p^\star)\le
 \eta+(1-\eta)
       (\epsilon_{\rm mass}+\epsilon_{\rm solve}).
 \label{eq:error}
\end{equation}
The bound separates representation support, probability assignment, and conditional solving.
\Cref{eq:bank} adds a separately measured sampling contribution in squared-error units. Exact-family
results report the TV terms and the bank contribution separately; they have different units and do
not form an additive attribution of an accuracy gap. Appendix~A states the
sufficient-representation assumptions and proves the bound.

\section{Experiments and Results}
\label{sec:experiments}
We test probability quality, reuse across questions, acquired-view revision, and end-to-end language
reasoning. Each comparison isolates a training objective, inference estimator, or observation
change.

\begin{table*}[t]
\centering\small\setlength{\tabcolsep}{3.0pt}
\caption{\textbf{Probability quality on RevealRooms.} All scene methods use $K_p=1{,}024$; soft heads retain their distributions. Entries are mean$_{\pm\mathrm{SD}}$ over five seeds. Bold and underlining mark the best and second-best displayed values, including ties. Extra optimization controls are evaluated on Natural (\textendash: not evaluated on Exact). Acc., ECE, and Drift are percentage quantities; xBrier is excess expected Brier over the true posterior.}
\label{tab:main}
\begin{tabular}{@{}lrrrrrr@{}}
\toprule
& \multicolumn{4}{c}{Natural} & \multicolumn{2}{c}{Exact} \\
\cmidrule(lr){2-5}\cmidrule(l){6-7}
Method & Acc.$\uparrow$ & Brier$\downarrow$ & NLL$\downarrow$ & ECE$\downarrow$ & xBrier$\downarrow$ & Drift$\downarrow$ \\
\midrule
Direct VLM & $75.26_{\scriptscriptstyle\pm 0.21}$ & $0.1705_{\scriptscriptstyle\pm 0.0011}$ & $0.519_{\scriptscriptstyle\pm 0.003}$ & $2.08_{\scriptscriptstyle\pm 0.13}$ & $0.0113_{\scriptscriptstyle\pm 0.0001}$ & $1.56_{\scriptscriptstyle\pm 0.01}$ \\
View Consensus & $76.16_{\scriptscriptstyle\pm 0.30}$ & $0.1640_{\scriptscriptstyle\pm 0.0016}$ & $0.502_{\scriptscriptstyle\pm 0.004}$ & $1.81_{\scriptscriptstyle\pm 0.20}$ & $0.0109_{\scriptscriptstyle\pm 0.0001}$ & $1.72_{\scriptscriptstyle\pm 0.01}$ \\
Scene Completion & $77.07_{\scriptscriptstyle\pm 0.11}$ & $0.1571_{\scriptscriptstyle\pm 0.0008}$ & $0.481_{\scriptscriptstyle\pm 0.002}$ & $0.95_{\scriptscriptstyle\pm 0.22}$ & $0.0101_{\scriptscriptstyle\pm 0.0001}$ & $1.41_{\scriptscriptstyle\pm 0.01}$ \\
Answer Distillation & $79.12_{\scriptscriptstyle\pm 0.22}$ & $0.1422_{\scriptscriptstyle\pm 0.0013}$ & $0.438_{\scriptscriptstyle\pm 0.004}$ & $0.85_{\scriptscriptstyle\pm 0.19}$ & $0.0092_{\scriptscriptstyle\pm 0.0001}$ & $1.35_{\scriptscriptstyle\pm 0.01}$ \\
Scene NLL & $79.22_{\scriptscriptstyle\pm 0.16}$ & $0.1424_{\scriptscriptstyle\pm 0.0012}$ & $0.439_{\scriptscriptstyle\pm 0.003}$ & $0.76_{\scriptscriptstyle\pm 0.08}$ & $0.0087_{\scriptscriptstyle\pm 0.0001}$ & $1.23_{\scriptscriptstyle\pm 0.01}$ \\
Scene NLL, View Subsets & $79.18_{\scriptscriptstyle\pm 0.13}$ & $0.1410_{\scriptscriptstyle\pm 0.0003}$ & $0.432_{\scriptscriptstyle\pm 0.001}$ & $0.75_{\scriptscriptstyle\pm 0.12}$ & \textendash & \textendash \\
Scene NLL, 40 Epochs & $79.71_{\scriptscriptstyle\pm 0.15}$ & $0.1392_{\scriptscriptstyle\pm 0.0012}$ & $0.429_{\scriptscriptstyle\pm 0.003}$ & $0.82_{\scriptscriptstyle\pm 0.09}$ & \textendash & \textendash \\
Scene NLL + Joint & $79.83_{\scriptscriptstyle\pm 0.32}$ & $0.1383_{\scriptscriptstyle\pm 0.0016}$ & $0.427_{\scriptscriptstyle\pm 0.004}$ & $\underline{0.68}_{\scriptscriptstyle\pm 0.15}$ & $0.0079_{\scriptscriptstyle\pm 0.0001}$ & $1.18_{\scriptscriptstyle\pm 0.01}$ \\
RevealBelief w/o Joint & $\underline{80.21}_{\scriptscriptstyle\pm 0.04}$ & $\underline{0.1346}_{\scriptscriptstyle\pm 0.0005}$ & $\underline{0.415}_{\scriptscriptstyle\pm 0.001}$ & $0.83_{\scriptscriptstyle\pm 0.09}$ & $\underline{0.0072}_{\scriptscriptstyle\pm 0.0001}$ & $\underline{0.98}_{\scriptscriptstyle\pm 0.01}$ \\
RevealBelief & $\mathbf{80.81}_{\scriptscriptstyle\pm 0.14}$ & $\mathbf{0.1313}_{\scriptscriptstyle\pm 0.0010}$ & $\mathbf{0.406}_{\scriptscriptstyle\pm 0.003}$ & $\mathbf{0.67}_{\scriptscriptstyle\pm 0.16}$ & $\mathbf{0.0065}_{\scriptscriptstyle\pm 0.0001}$ & $\mathbf{0.89}_{\scriptscriptstyle\pm 0.01}$ \\
\bottomrule
\end{tabular}
\end{table*}

\subsection{Setup}
\paragraph{RevealRooms.}
Natural uses ProcTHOR layouts~\cite{procthor}: 150,000 training and 5,000 each selection,
calibration, and test episodes, grouped by base layout. Students receive one, two, or four views;
teachers receive four extra training views. Five query families give 20,000 binary test outcomes.
Exact has 40,000 training families, including 2,000 for selection, plus 2,000 calibration and 4,000
test families. Four arrangements per family share an identical initial image; nonuniform priors
provide known event probabilities. Appendix~C specifies the generators and balanced
control.

\paragraph{Real Scans.}
ScanNet v2~\cite{scannet} uses 1,001 fitting, 100 selection, and 100 calibration scans, grouped into
469, 45, and 51 disjoint physical spaces. The 312 official validation scans span 142 other spaces.
Sixteen sets per scan and four questions per set provide 19,968 outcomes. Starting sets contain two,
four, or eight posed RGB frames, with a one-frame acquisition condition. Reference-supported regions
determine labels before question sampling; test inventory is inferred from images.
Appendix~C.8 reports estimated-pose and second-dataset transfer.

\paragraph{Controls.}
All methods share inputs, splits, and calibration access. Scene arms share architecture, labels, and
execution with $K_p=1{,}024$. \emph{Direct VLM} fits a Qwen3-VL adapter on partial-view answers.
\emph{View Consensus} pools imagined-view tokens in one decoder without source-hypothesis
identifiers. \emph{Scene Completion} uses a conditional latent-variable completion objective on our
state vocabulary, following the scene-belief motivation of 3D-Belief~\cite{belief3d}. \emph{Answer
Distillation} receives the same eight teacher samples and trains marginal and joint answer heads.
Scene NLL and distillation are crossed with the same joint loss. Extra controls match the teacher's
view-subset schedule or double Scene NLL to 40 epochs. Each arm fits its temperature on the same
split.

\paragraph{Scores and Uncertainty.}
We report accuracy, binary Brier, half-count log loss, ten-bin confidence ECE, and AURC. Exact also
measures excess Brier, confidence gap, and irrelevant-view drift. Shared-protocol comparisons report
five-seed means and SDs; differences and relative changes use unrounded scores. Paired intervals use
2,000 crossed resamples of seeds and complete families or physical spaces, retaining related
questions and repeat scans. Appendix~C.4 defines the estimand, score regularization, and
tie handling.

\subsection{Main Results}
\paragraph{Natural Scenes.}
RevealBelief reduces Brier to $0.1313$, against $0.1424$ for Scene NLL and $0.1422$ for
answer distillation (\cref{tab:main}). Its log loss is $0.406$, and accuracy is $80.81\%$. The gain
remains $0.0078$ against the 40-epoch Scene NLL control. View-subset training improves the
likelihood baseline, while revealing-view distillation and joint supervision produce a further gain.
The paired Brier improvement over standard Scene NLL is $0.0110$, with 95\% interval
$[0.0100,0.0121]$.

\paragraph{Exact Posterior.}
Excess Brier falls from $0.0135$ at $K=32$ to $0.0065$ at $K_p=1{,}024$. At the larger
budget, RevealBelief improves over Scene NLL ($0.0087$) and answer distillation ($0.0092$). On
Natural, Brier at $K=32,128,1{,}024$ is $0.1353$, $0.1323$, and $0.1313$. On Exact, corrected $K=32$
excess Brier is $0.00628$; direct family integration gives $0.00625$. The agreement identifies the
sampling contribution at a fixed belief. Appendix~C.5 reports all budgets and the
measured TV components.

\begin{table*}[t]
\centering\footnotesize\setlength{\tabcolsep}{1.6pt}
\caption{\textbf{ScanNet outcome calibration and acquired-view revision.} All shared methods from \cref{tab:main} use the same observations, calibration scenes, and scoring rules. $R_{25}$ and $R_{50}$ are selective errors at 25\% and 50\% coverage. The right block adds one acquired frame and refreshes predictions. Entries are mean$_{\pm\mathrm{SD}}$; accuracy, ECE, AURC, and selective errors are in percent. The subset control isolates its initial-input training schedule.}
\label{tab:real}
\begin{tabular}{@{}lrrrrrrrrr@{}}
\toprule
& \multicolumn{7}{c}{Initial acquired set} & \multicolumn{2}{c}{Added acquisition} \\
\cmidrule(lr){2-8}\cmidrule(l){9-10}
Method & Acc.$\uparrow$ & Brier$\downarrow$ & NLL$\downarrow$ & ECE$\downarrow$ & AURC$\downarrow$ & $R_{25}\!\downarrow$ & $R_{50}\!\downarrow$ & Acc.$\uparrow$ & Brier$\downarrow$ \\
\midrule
Direct VLM & $71.65_{\scriptscriptstyle\pm 0.31}$ & $0.1901_{\scriptscriptstyle\pm 0.0012}$ & $0.563_{\scriptscriptstyle\pm 0.003}$ & $2.04_{\scriptscriptstyle\pm 0.12}$ & $18.31_{\scriptscriptstyle\pm 0.25}$ & $13.95_{\scriptscriptstyle\pm 0.39}$ & $18.52_{\scriptscriptstyle\pm 0.23}$ & $78.40_{\scriptscriptstyle\pm 0.28}$ & $0.1531_{\scriptscriptstyle\pm 0.0012}$ \\
View Consensus & $72.67_{\scriptscriptstyle\pm 0.23}$ & $0.1838_{\scriptscriptstyle\pm 0.0015}$ & $0.548_{\scriptscriptstyle\pm 0.004}$ & $1.55_{\scriptscriptstyle\pm 0.18}$ & $16.77_{\scriptscriptstyle\pm 0.34}$ & $12.14_{\scriptscriptstyle\pm 0.42}$ & $16.83_{\scriptscriptstyle\pm 0.54}$ & $79.55_{\scriptscriptstyle\pm 0.29}$ & $0.1451_{\scriptscriptstyle\pm 0.0015}$ \\
Scene Completion & $72.97_{\scriptscriptstyle\pm 0.19}$ & $0.1803_{\scriptscriptstyle\pm 0.0009}$ & $0.537_{\scriptscriptstyle\pm 0.002}$ & $\mathbf{0.67}_{\scriptscriptstyle\pm 0.16}$ & $15.59_{\scriptscriptstyle\pm 0.18}$ & $9.98_{\scriptscriptstyle\pm 0.42}$ & $16.09_{\scriptscriptstyle\pm 0.29}$ & $79.92_{\scriptscriptstyle\pm 0.12}$ & $0.1400_{\scriptscriptstyle\pm 0.0008}$ \\
Answer Distillation & $74.41_{\scriptscriptstyle\pm 0.33}$ & $0.1706_{\scriptscriptstyle\pm 0.0013}$ & $0.510_{\scriptscriptstyle\pm 0.003}$ & $0.86_{\scriptscriptstyle\pm 0.16}$ & $13.22_{\scriptscriptstyle\pm 0.23}$ & $6.83_{\scriptscriptstyle\pm 0.30}$ & $13.55_{\scriptscriptstyle\pm 0.15}$ & $81.30_{\scriptscriptstyle\pm 0.23}$ & $0.1293_{\scriptscriptstyle\pm 0.0011}$ \\
Scene NLL & $74.69_{\scriptscriptstyle\pm 0.20}$ & $0.1689_{\scriptscriptstyle\pm 0.0009}$ & $0.506_{\scriptscriptstyle\pm 0.002}$ & $0.74_{\scriptscriptstyle\pm 0.15}$ & $12.96_{\scriptscriptstyle\pm 0.19}$ & $6.68_{\scriptscriptstyle\pm 0.19}$ & $13.16_{\scriptscriptstyle\pm 0.22}$ & $81.94_{\scriptscriptstyle\pm 0.13}$ & $0.1262_{\scriptscriptstyle\pm 0.0009}$ \\
Scene NLL, View Subsets & $74.55_{\scriptscriptstyle\pm 0.11}$ & $0.1677_{\scriptscriptstyle\pm 0.0003}$ & $0.502_{\scriptscriptstyle\pm 0.001}$ & $\underline{0.68}_{\scriptscriptstyle\pm 0.20}$ & $12.55_{\scriptscriptstyle\pm 0.06}$ & $6.00_{\scriptscriptstyle\pm 0.16}$ & $12.57_{\scriptscriptstyle\pm 0.10}$ & \textendash & \textendash \\
Scene NLL + Joint & $75.19_{\scriptscriptstyle\pm 0.21}$ & $0.1653_{\scriptscriptstyle\pm 0.0015}$ & $0.496_{\scriptscriptstyle\pm 0.004}$ & $0.99_{\scriptscriptstyle\pm 0.09}$ & $12.18_{\scriptscriptstyle\pm 0.25}$ & $5.73_{\scriptscriptstyle\pm 0.29}$ & $12.25_{\scriptscriptstyle\pm 0.39}$ & $82.78_{\scriptscriptstyle\pm 0.21}$ & $0.1199_{\scriptscriptstyle\pm 0.0012}$ \\
RevealBelief w/o Joint & $\underline{75.93}_{\scriptscriptstyle\pm 0.07}$ & $\underline{0.1605}_{\scriptscriptstyle\pm 0.0006}$ & $\underline{0.483}_{\scriptscriptstyle\pm 0.001}$ & $0.82_{\scriptscriptstyle\pm 0.10}$ & $\underline{11.30}_{\scriptscriptstyle\pm 0.09}$ & $\underline{4.76}_{\scriptscriptstyle\pm 0.08}$ & $\underline{11.06}_{\scriptscriptstyle\pm 0.19}$ & $\underline{83.60}_{\scriptscriptstyle\pm 0.08}$ & $\underline{0.1142}_{\scriptscriptstyle\pm 0.0004}$ \\
RevealBelief & $\mathbf{76.40}_{\scriptscriptstyle\pm 0.20}$ & $\mathbf{0.1574}_{\scriptscriptstyle\pm 0.0012}$ & $\mathbf{0.475}_{\scriptscriptstyle\pm 0.003}$ & $0.93_{\scriptscriptstyle\pm 0.19}$ & $\mathbf{10.79}_{\scriptscriptstyle\pm 0.19}$ & $\mathbf{4.40}_{\scriptscriptstyle\pm 0.27}$ & $\mathbf{10.44}_{\scriptscriptstyle\pm 0.32}$ & $\mathbf{84.45}_{\scriptscriptstyle\pm 0.25}$ & $\mathbf{0.1087}_{\scriptscriptstyle\pm 0.0010}$ \\
\bottomrule
\end{tabular}
\end{table*}

\paragraph{Real-Scene Calibration.}
The Brier and selective-risk gains carry to ScanNet. RevealBelief lowers Brier by $6.8\%$
and AURC by $16.8\%$ relative to Scene NLL (\cref{tab:real}). Its log loss is $0.475$ and accuracy
$76.40\%$. At 25\% coverage its error is $4.40\%$, against $6.68\%$ for Scene NLL; at 50\% coverage
the corresponding errors are $10.44\%$ and $13.16\%$. ECE is below one percentage point. These
scores connect probability quality to useful answer selection.

\paragraph{Acquired Evidence.}
One acquired view increases the Brier advantage to $13.9\%$. RevealBelief reaches $84.45\%$
accuracy and Brier $0.1087$, making $13.9\%$ fewer errors than Scene NLL. All methods receive the
same pose-diverse frame, selected from unused scan frames by a fixed max-min camera-distance rule
before questions are sampled (Appendix~C.1). The observation re-encodes the scene and
refreshes every related prediction.

\paragraph{Visibility Before and After Acquisition.}
Both visibility groups improve; the absolute advantage is larger for initially visible
targets. Initial Brier gains over Scene NLL are 0.0139 for visible and 0.0090 for occluded targets.
After acquisition, RevealBelief changes from 0.1248 to 0.0691 in the visible group and from 0.1895
to 0.1475 in the occluded group. The corresponding Scene NLL scores are 0.0882 and 0.1635, giving
advantages of 0.0191 and 0.0160. Appendix~C.7 reports the paired groups and predicate
breakdown.

\begin{figure*}[t]
\centering
\includegraphics[width=\textwidth]{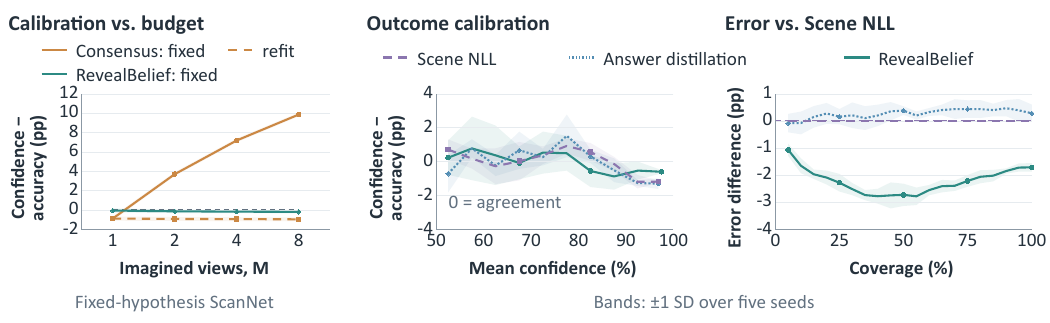}
\caption{\textbf{Computation budget and useful confidence.} \textbf{Left:} signed confidence gap on
  the fixed-hypothesis ScanNet sensitivity cohort, with either the $M=1$ temperature or a separately
  fitted budget-specific temperature. \textbf{Middle:} binned confidence minus accuracy on the main
  $K_p=1{,}024$ ScanNet comparison. \textbf{Right:} selective-error difference to Scene NLL across
  retained coverage; negative values favor the plotted method. Lines average five seeds; bands show
  paired one-standard-deviation variation. The budget cohort fixes the scene/readout profile;
  \cref{tab:readout} separately measures language improvements from additional views.}
\label{fig:diagnostics}
\end{figure*}

\subsection{Ablations and Question Reuse}
\label{sec:ablations}
\begin{table}[t]
\centering\footnotesize\setlength{\tabcolsep}{2.0pt}
\caption{\textbf{Objectives and five-family reuse.} Standard-checkpoint Natural joint log loss is followed by the macro-average across five separately trained leave-one-family-out settings. Joint log loss scores the four feasible three-answer signatures specified in Appendix~C.1. All entries are mean$_{\pm\mathrm{SD}}$.}
\label{tab:ablation}
\begin{tabular}{@{}lrrr@{}}
\toprule
& Standard & \multicolumn{2}{c}{Held-out families} \\
\cmidrule(lr){2-2}\cmidrule(l){3-4}
Method & Joint NLL$\downarrow$ & Acc.$\uparrow$ & Brier$\downarrow$ \\
\midrule
Answer Distill. & $0.601_{\scriptscriptstyle\pm 0.005}$ & $71.69_{\scriptscriptstyle\pm 0.24}$ & $0.1853_{\scriptscriptstyle\pm 0.0011}$ \\
Scene NLL & $0.588_{\scriptscriptstyle\pm 0.007}$ & $74.10_{\scriptscriptstyle\pm 0.34}$ & $0.1731_{\scriptscriptstyle\pm 0.0015}$ \\
Scene NLL + Joint & $\underline{0.496}_{\scriptscriptstyle\pm 0.007}$ & $74.73_{\scriptscriptstyle\pm 0.27}$ & $0.1691_{\scriptscriptstyle\pm 0.0017}$ \\
RevealBelief w/o Joint & $0.516_{\scriptscriptstyle\pm 0.010}$ & $\underline{75.16}_{\scriptscriptstyle\pm 0.06}$ & $\underline{0.1663}_{\scriptscriptstyle\pm 0.0006}$ \\
RevealBelief & $\mathbf{0.421}_{\scriptscriptstyle\pm 0.007}$ & $\mathbf{76.26}_{\scriptscriptstyle\pm 0.23}$ & $\mathbf{0.1598}_{\scriptscriptstyle\pm 0.0011}$ \\
\bottomrule
\end{tabular}
\end{table}

\paragraph{Teacher and Joint Supervision.}
Each objective improves the shared representation. Teacher distillation without the joint
loss gives Natural Brier $0.1346$. The joint term improves Scene NLL by $0.0040$ and distillation by
$0.0032$; their combination achieves the best joint log loss (\cref{tab:ablation}). On ScanNet the
four-extra-view teacher has Brier $0.1127$, compared with $0.1677$ from the same checkpoint at zero
extra views. A separately optimized zero-extra-view student and a label-corruption control appear in
Appendix~C.3.

\paragraph{Unseen Questions.}
The advantage holds across all five excluded families. Each leave-one-family-out run
removes its operator from every task-specific answer target, joint cluster, solver-fitting example,
and answer-calibration set. Scene supervision remains unchanged. RevealBelief averages $76.26\%$
accuracy and Brier $0.1598$, improving accuracy by $4.6$ points over answer distillation and $2.2$
over Scene NLL. Across families its accuracy ranges from $75.80$ to $77.05\%$. The answer-head
comparison measures reuse; the Scene NLL comparison isolates supervision.

\paragraph{Data and Optimization.}
The gain persists with doubled student updates and less training data. \cref{tab:main}
includes the 40-epoch Scene NLL and random-view-subset controls. The doubled-update control costs
1,568 GPU-hours, compared with 2,960 for our teacher and student combined; both exclude the shared
codebook. At 10\% of Natural training families, Brier is $0.1453$ for RevealBelief, $0.1567$ for
Scene NLL, and $0.1565$ for answer distillation. Appendix Table~A8 reports all three
methods at 10, 25, 50, and 100\% with the teacher trained on the same restricted subset. The
exact-teacher identity establishes the common population target; these comparisons measure
finite-model learning.

\subsection{Imagined Views}
\begin{table}[t]
\centering\footnotesize\setlength{\tabcolsep}{1.2pt}
\caption{\textbf{End-to-end composed-language evaluation.} Natural has 8,000 queries and ScanNet 7,984. Ground truth is the actual scene answer. The two RevealBelief rows share the same $32$ hypotheses and frozen solver; the final prediction averages soft conditional scores. Four views are rendered per hypothesis. Brier is multiplied by 100. Entries are mean$_{\pm\mathrm{SD}}$.}
\label{tab:readout}
\begin{tabular}{@{}lrrrr@{}}
\toprule
& \multicolumn{2}{c}{Natural} & \multicolumn{2}{c}{ScanNet} \\
\cmidrule(lr){2-3}\cmidrule(l){4-5}
Method & Acc.$\uparrow$ & Brier$\times100$ & Acc.$\uparrow$ & Brier$\times100$ \\
\midrule
Direct VLM & $74.56_{\scriptscriptstyle\pm 0.27}$ & $17.38_{\scriptscriptstyle\pm 0.14}$ & $71.31_{\scriptscriptstyle\pm 0.24}$ & $19.06_{\scriptscriptstyle\pm 0.10}$ \\
Answer Distillation & $78.53_{\scriptscriptstyle\pm 0.18}$ & $14.52_{\scriptscriptstyle\pm 0.11}$ & $74.24_{\scriptscriptstyle\pm 0.27}$ & $17.12_{\scriptscriptstyle\pm 0.14}$ \\
RevealBelief, Code & $\underline{79.68}_{\scriptscriptstyle\pm 0.25}$ & $\underline{14.33}_{\scriptscriptstyle\pm 0.06}$ & $\underline{75.23}_{\scriptscriptstyle\pm 0.36}$ & $\underline{16.81}_{\scriptscriptstyle\pm 0.19}$ \\
RevealBelief, Views & $\mathbf{79.77}_{\scriptscriptstyle\pm 0.28}$ & $\mathbf{13.98}_{\scriptscriptstyle\pm 0.06}$ & $\mathbf{75.36}_{\scriptscriptstyle\pm 0.24}$ & $\mathbf{16.51}_{\scriptscriptstyle\pm 0.18}$ \\
\bottomrule
\end{tabular}
\end{table}

\paragraph{End-to-End Language Reasoning.}
Rendered hypotheses improve answers about the actual scene. \cref{tab:readout} evaluates
composed-language questions on Natural and ScanNet, without supplying their programs. The final
prediction averages the solver's soft scores over the $32$ render roots. Four views reduce Natural
Brier from $0.1433$ to $0.1398$ and ScanNet Brier from $0.1681$ to $0.1651$, with accuracies of
$79.77\%$ and $75.36\%$. Brier uses the observed scene's outcome rather than each candidate's
internal truth. The candidate-conditional study in Appendix~C.6 separately attributes a
$2.9$-point gain to reading the same hypotheses more accurately.

\paragraph{Budget Sensitivity.}
Four views capture nearly all of the conditional-readout gain. Eight add $0.2$ points on
the candidate cohort. \Cref{fig:diagnostics} includes both fixed and budget-specific calibration for
View Consensus. Recalibration removes much of its confidence drift; RevealBelief keeps a small drift
with the fixed calibration in this diagnostic. The scene weights are invariant to computational
copies, while probabilities can change when additional views improve the conditional solver. One
bank caches 128 renders for an eight-question batch, amortizing rendering to 16 images per question.
Sampling the probability bank is charged separately from these renders.

\paragraph{Inference Cost.}
Probability sampling dominates the cold geometric path. On eight H100 80\,GB GPUs, sampling
$1{,}024$ states takes 43.6\,s per observation set; the full language path takes 55.94\,s for eight
questions. A $128$-state executor reaches Natural Brier $0.1323$ versus $0.1313$ at $1{,}024$
(Appendix Table~A11), using one eighth as many hypotheses. Cached geometric questions take
3.4\,ms each. Acquiring a new image refreshes the states and render cache (Appendix~C.10).

\begin{figure*}[t]
\centering
\includegraphics[width=\textwidth]{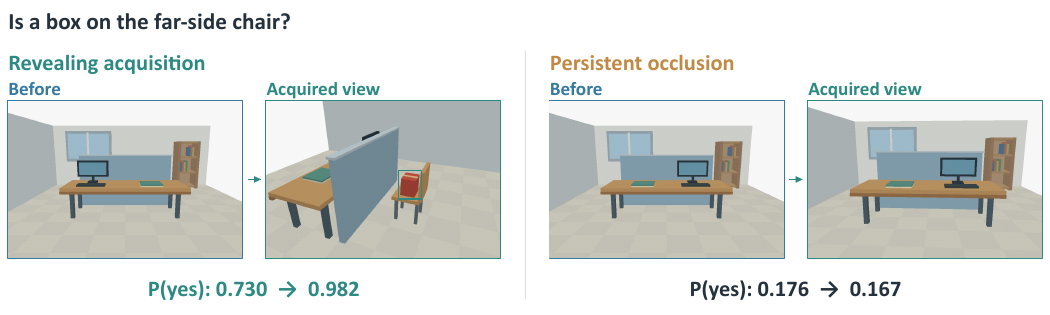}
\caption{\textbf{Revealing acquisition and persistent occlusion.} \textbf{Left:} the acquired view
  exposes the target and raises occupied-event mass from $748/1{,}024$ to $1{,}006/1{,}024$.
  \textbf{Right:} continued occlusion leaves the occupied family underweighted ($180/1{,}024$ to
  $171/1{,}024$; reference probability $0.72$). Views share a coordinate frame within each
  RevealRooms case. The small right-hand change reflects fresh sampling; further renders of existing
  hypotheses cannot repair an incorrect prior.}
\label{fig:qualitative}
\end{figure*}

\subsection{Transfer and Qualitative Behavior}
\begin{table}[t]
\centering\footnotesize\setlength{\tabcolsep}{3.0pt}
\caption{\textbf{VSI-Bench video transfer.} Every arm uses 16 uniformly spaced frames, the same VGGT frontend, and a Qwen3-VL readout. MCA averages the four multiple-choice task groups; it is distinct from the full eight-task score. Entries are accuracy percentages from source-selected checkpoints.}
\label{tab:public}
\begin{tabular}{@{}lrr@{}}
\toprule
Method & 8B MCA & 32B MCA \\
\midrule
Direct + Self-Consistency & 44.1 & 50.2 \\
Scene Completion & 46.8 & 52.4 \\
Scene NLL & 47.6 & 53.2 \\
RevealBelief w/o Joint & \underline{48.6} & \underline{53.8} \\
RevealBelief & \textbf{49.2} & \textbf{54.3} \\
\bottomrule
\end{tabular}
\end{table}

\paragraph{Public Benchmarks.}
The scene interface transfers to video reasoning. On VSI-Bench~\cite{vsibench},
RevealBelief reaches 49.2\% four-task multiple-choice accuracy with the 8B solver and 54.3\% with
32B (\cref{tab:public}). All transfer arms share 16 uniformly spaced frames, VGGT geometry, and the
observed trajectory; target labels do not enter fitting or calibration. Appendix~C.9
provides the per-task breakdown, published SAT-Real context~\cite{avic}, and the distinct abstention
protocol of SpatialUncertain~\cite{spatialuncertain}.

\paragraph{Pose and Dataset Transfer.}
The ordering persists when the camera frontend changes. Replacing provided ScanNet poses
with VGGT estimates gives Brier $0.1592$ for RevealBelief and $0.1707$ for Scene NLL. Zero-shot
ScanNet++~\cite{scannetpp} transfer uses its 50-scene validation partition, mapped to the shared
category vocabulary, and retains the source calibration. Appendix~C.8 provides accuracy,
Brier, and the full protocol.

\paragraph{Qualitative Mechanisms.}
The acquired image determines which ambiguity can be resolved. In \Cref{fig:qualitative},
the revealing view distinguishes hypotheses that explain the initial pixels, concentrating
probability on the occupied event. Re-encoding also permits hypotheses absent from the previous
finite bank to enter the prediction. Persistent occlusion leaves the event unresolved and exposes an
error in the learned family mass. Additional renders can clarify each proposed scene, while the
relative probability of those scenes must come from the belief. These cases connect the observed
update to the information in the acquisition; Appendix~C.5 separately quantifies mass
and sampling errors.

\section{Limitations}
\label{sec:discussion}
The scene belief covers the trained object vocabulary and static geometry. Dynamic objects,
open-vocabulary inventories, and learned acquisition policies are natural extensions. A new view can
leave a target occluded (\cref{fig:qualitative}). Reference-complete regions and camera quality
determine which geometric outcomes can be scored reliably; the estimated-pose experiment measures
one practical source of error. The teacher and student can share misspecification, and an unknown
shift in hidden-scene frequencies cannot be identified from unchanged acquired images. The large
bank improves Monte Carlo precision, while broader support and better learned mass require changes
to the representation or training signal.

\section{Conclusion}
RevealBelief learns a reusable scene distribution from revealing views, enabling consistent answers to related queries through joint supervision. By separating probability estimation from imagined-view computation, it supports reasoning under partial observability and coherent updates from new evidence. On ScanNet, the Brier advantage over scene likelihood increases from $6.8\%$ to $13.9\%$ after an acquired view. Moreover, five-family reuse improves accuracy over answer distillation by $4.6$ percentage points, highlighting the benefit of learning a shared scene representation. These results support calibrated probabilistic reasoning, coherent belief updates, and efficient reuse across diverse query families.

\clearpage


{
    \small
    \bibliographystyle{ieeenat_fullname}
    \bibliography{main}
}


\clearpage
\appendix
\floatstyle{ruled}
\newfloat{algorithm}{tbp}{loa}
\floatname{algorithm}{Algorithm}
\newtheorem{appendixproposition}{Proposition}[section]
\setlist[itemize]{leftmargin=*,nosep}
\setlist[enumerate]{leftmargin=*,nosep}
\graphicspath{{figures/}}
\crefalias{section}{appendix}
\crefalias{subsection}{appendix}
\renewcommand{\thefigure}{A\arabic{figure}}
\renewcommand{\thetable}{A\arabic{table}}
\renewcommand{\thealgorithm}{A\arabic{algorithm}}
\numberwithin{equation}{section}
\label{app:start}
\section{Assumptions and Theoretical Properties}
\label{sec:appendix-theory}
\label{app:theory}

\subsection{Scope And Notation}
Let \(\mathcal Z\) be the finite space of categorical scene codes. The general model predicts
presence, category, geometry, shape, and appearance within a fixed slot capacity. The four-object
permutation reference is a separate space with a fixed inventory. We analyze one fixed
data-generating law for acquired observations, additional views, and reference scene states.
Real-scan statements concern the reference-supported geometry specified by the evaluation protocol.

A representation is sufficient for a query family when its answer kernel \(f_z\) satisfies \(P(Y\mid
Z,\mathcal{O},q)=f_Z(Y\mid q)\). Deterministic execution requires quantization and geometry
tolerances to preserve the scored relation. Question selection carries no extra hidden-state
information, \(Q\perp Z\mid\mathcal{O}\). Teacher-derived joint targets additionally use clusters
sampled independently of \((Z,\mathcal V)\) given \(\mathcal{O}\), as enforced by the fixed-episode
construction. When language supplies a scene fact, that fact instead belongs to the conditioning
observation. These assumptions identify the target distribution; they do not assert that a trained
teacher, student, camera estimator, or solver equals that target. Learned-model approximation and
unknown prior shift are treated separately below.

\begin{table}[t]
\centering\small
\begin{tabular}{@{}p{.25\columnwidth}p{.69\columnwidth}@{}}
\toprule
Symbol & Definition\\
\midrule
\(Z,\mathcal Z\) & Scene state and finite state space.\\
\(\mathcal{O},\mathcal V\) & Acquired set and additional training views.\\
\(P_o\) & True posterior \(P(Z\mid\mathcal{O}=o)\).\\
\(T_\psi,b_\phi\) & Multi-view teacher and partial-view student.\\
\(L\) & Length of the serialized scene code.\\
\(K,M,m\) & Hypotheses, views per hypothesis, questions per cluster.\\
\(z_k,w_k,u_k\) & Hypothesis, normalized weight, immutable identifier.\\
\(\mathcal S,\eta\) & Represented support and true mass outside it.\\
\(Q,F_Q(z)\) & Question cluster and answer signature.\\
\(\widehat\pi\) & Predicted mass on represented support.\\
\(\operatorname{TV}(P,Q)\) & \(\frac12\sum_z|P(z)-Q(z)|\).\\
\bottomrule
\end{tabular}
\caption{Notation for the theoretical and implementation details.}
\label{tab:appendix-notation}
\end{table}

\subsection{Distillation From Additional Observations}
For any normalized teacher, define its observation-conditional average:
\begin{equation}
 \bar T_o(z)=
 \mathbb{E}_{\mathcal V\mid o}[T_\psi(z\mid o,\mathcal V)].
 \label{eq:app-average-teacher}
\end{equation}
The population belief loss is:
\begin{equation}
 \mathcal L_{\rm belief}
 =\mathbb{E}_{\mathcal{O},\mathcal V}
 \left[-\sum_z T_\psi(z\mid\mathcal{O},\mathcal V)
               \log b_\phi(z\mid\mathcal{O})\right].
 \label{eq:app-distillation}
\end{equation}

\begin{appendixproposition}[Revealing-View Distillation]
\label{prop:app-distillation}
Over unrestricted conditional distributions, a population minimizer of \cref{eq:app-distillation}
satisfies \(b^\star(\cdot\mid o)=\bar T_o\) almost surely. If \(T_\psi(\cdot\mid o,v)=P(Z\mid o,v)\)
almost surely, then \(b^\star(\cdot\mid o)=P_o\).
\end{appendixproposition}

\begin{proof}
Condition on \(\mathcal{O}=o\). A finite sum can be interchanged with the view expectation. The
conditional risk is:
\begin{equation}
 \begin{aligned}
 \mathcal L_o(b)
 &=-\sum_z\bar T_o(z)\log b(z\mid o)\\
 &=H(\bar T_o)+\operatorname{KL}(\bar T_o\|b(\cdot\mid o)).
 \end{aligned}
 \label{eq:app-distillation-kl}
\end{equation}
Nonnegativity of KL proves the first claim, with infinite loss for zero student probability on
positive teacher mass. For an exact teacher, the tower property gives:
\begin{equation}
 \begin{aligned}
 \bar T_o(z)
 &=\mathbb{E}_{\mathcal V\mid o}[P(Z=z\mid o,\mathcal V)]\\
 &=P(Z=z\mid o).
 \end{aligned}
\end{equation}
This proves the second claim.
\end{proof}

The proposition permits revealing views that leave residual ambiguity. A realized scene remains a
valid one-hot supervised label for NLL; this does not assert that its partial-view predictive
distribution is a point mass. Unrestricted population minimizers provide the reference target for
finite-model estimation.

\paragraph{Imperfect Teachers.}
For fixed \(o\), abbreviate the teacher and true revealed posterior by \(T_v\) and \(P_v\).
Convexity of absolute value gives:
\begin{equation}
 \begin{aligned}
 \operatorname{TV}(\bar T_o,P_o)
 &=\frac12\sum_z
   \left|\mathbb{E}_{\mathcal V\mid o}
       [T_{\mathcal V}(z)-P_{\mathcal V}(z)]\right|\\
 &\le \mathbb{E}_{\mathcal V\mid o}
           \operatorname{TV}(T_{\mathcal V},P_{\mathcal V}).
 \end{aligned}
 \label{eq:app-teacher-tv}
\end{equation}
For a fitted student, the triangle inequality therefore yields:
\begin{equation}
 \begin{aligned}
 \operatorname{TV}(b_\phi(\cdot\mid o),P_o)
 \le{}&\operatorname{TV}(b_\phi(\cdot\mid o),\bar T_o)\\
 &+\mathbb{E}_{\mathcal V\mid o}
       \operatorname{TV}(T_{\mathcal V},P_{\mathcal V}).
 \end{aligned}
 \label{eq:app-student-tv}
\end{equation}
Student approximation and average teacher error enter separately. Outcome-based proper scores can
assess learned predictions on real scenes without making the unknown conditional TV terms directly
observable.

\subsection{Conditional Gradient Averaging}
Fix the student parameters and let:
\begin{equation}
 g(Z,\mathcal{O})=-\nabla_\phi\log b_\phi(Z\mid\mathcal{O}).
\end{equation}
Assume finite second moments and fresh population episodes. Teacher parameters are held fixed when
differentiating the student objective.

\begin{appendixproposition}[Conditional Gradient Averaging]
\label{prop:app-gradient-variance}
For an exact teacher, define \(g_T=\mathbb{E}[g(Z,\mathcal{O})\mid\mathcal{O},\mathcal V]\). Then:
\begin{equation}
 \mathbb{E}[g_T]=\mathbb{E}[g],\qquad
 \operatorname{Cov}(g_T)\preceq\operatorname{Cov}(g).
 \label{eq:gradientvariance}
\end{equation}
For \(J\) conditionally independent teacher samples, their mean gradient \(\bar g_J\) has covariance:
\begin{equation}
 \begin{aligned}
 \operatorname{Cov}(\bar g_J)
 ={}&\operatorname{Cov}(g_T)\\
 &+\frac1J\mathbb{E}[
     \operatorname{Cov}(g\mid\mathcal{O},\mathcal V)].
 \end{aligned}
 \label{eq:app-gradient-variance}
\end{equation}
\end{appendixproposition}

\begin{proof}
The exact teacher samples the conditional law of \(Z\); differentiating its cross-entropy gives the
conditional average \(g_T\). The tower property proves equality of means. Total covariance gives:
\begin{equation}
 \begin{aligned}
 \operatorname{Cov}(g)
 ={}&\operatorname{Cov}(g_T)\\
 &+\mathbb{E}[\operatorname{Cov}(g\mid\mathcal{O},\mathcal V)].
 \end{aligned}
\end{equation}
The second term is positive semidefinite. Conditional independence divides the conditional
covariance of the sample mean by \(J\). Applying total covariance again gives
\cref{eq:app-gradient-variance}.
\end{proof}

This is the classical conditional-averaging identity. With \(J=1\), its covariance equals that of
the point-label gradient. A fully identifying reveal makes the conditional covariance zero, so all
these estimators coincide. Increasing the revealed information can reduce the uncertainty remaining
to average, rather than reducing \(\operatorname{Cov}(g_T)\) itself. A learned teacher can change
the mean gradient, so the covariance ordering is not transferred to it by architecture or
normalization. The identity explains a source of sampling variance, not an optimization or
generalization guarantee.

\paragraph{Known Family Alternatives.}
The Exact reference supplies training-only family metadata \(F\), with compatible states and prior
probabilities. Common conditioning defines:
\begin{equation}
 \begin{aligned}
 \mu_F(z\mid o)&=P(Z=z\mid o,F),\\
 T_F(z\mid o,v)&=P(Z=z\mid o,v,F).
 \end{aligned}
\end{equation}
Family-averaged scene NLL is:
\begin{equation}
 \mathcal L_F(o,F)
 =-\sum_z\mu_F(z\mid o)\log b_\phi(z\mid o).
 \label{eq:app-family-nll}
\end{equation}
The tower property implies:
\begin{equation}
 \mathbb{E}_{\mathcal V\mid o,F}
 [\mathcal L_{\rm belief}(o,\mathcal V,F)]
 =\mathcal L_F(o,F).
 \label{eq:app-family-identity}
\end{equation}
Complete averaging over revealing outcomes and teacher states therefore reproduces the
family-averaged loss. The identity also holds for the expected joint proper score, so adding the
same joint objective preserves parity. This averages log losses, not the logarithm of an average
predicted mass.

Let \(g_F=\mathbb{E}[g\mid\mathcal{O},F]\) and \(g_{T,F}=\mathbb{E}[g\mid\mathcal{O},\mathcal
V,F]\). Nested conditioning and total covariance give:
\begin{equation}
 \operatorname{Cov}(g_F)
 \preceq\operatorname{Cov}(g_{T,F})
 \preceq\operatorname{Cov}(g).
 \label{eq:app-family-variance}
\end{equation}
The common \(F\) is essential. If \(F\) is not determined by \(\mathcal{O}\), averaging \(\mu_F\)
over \(F\mid\mathcal{O}\) restores \(P(Z\mid\mathcal{O})\). Family metadata constructs supervision
and does not enter the student's test input. In a nonenumerable space, a learned teacher supplies an
approximate target without an explicit family sum; nonenumerability alone establishes neither its
necessity nor superiority over Scene NLL.

For a four-state calculation, take \(P=(.55,.10,.10,.25)\), \(b_\phi=(.4,.2,.2,.2)\), and
categorical-logit gradient \(g=b_\phi-e_Z\). A partial reveal identifies either \(\{1,2\}\) or
\(\{3,4\}\). \cref{tab:app-gradient-variance} gives the resulting covariance traces; every column
has the same mean gradient. The zero-variance no-revelation oracle presupposes that the true prior
is already known.

\begin{table}[t]
\centering\small
\setlength{\tabcolsep}{4pt}
\begin{tabular}{@{}lrrr@{}}
\toprule
Revelation & Point NLL & Exact teacher & \(J=8\)\\
\midrule
None & 0.615000 & 0.000000 & 0.076875\\
One predicate & 0.615000 & 0.302912 & 0.341923\\
Full & 0.615000 & 0.615000 & 0.615000\\
\bottomrule
\end{tabular}
\caption{Trace of gradient covariance in the four-state reference. Conditional averaging changes variance while preserving the expected gradient.}
\label{tab:app-gradient-variance}
\end{table}

\subsection{Answer Error And Represented Support}
Fix an observation and question. Let \(\mathcal S\subseteq\mathcal Z\) be the represented states,
\(\eta=P_o(\mathcal S^c)\), and \(P_{\mathcal S}=P_o(\cdot\mid\mathcal S)\) for \(\eta<1\). Repeated
independent samples of one state are aggregated by summing their weights. Write \(\widehat\pi\) for
the predicted mass on \(\mathcal S\), and \(p_z\) for the solver's answer kernel. Define:
\begin{equation}
 \begin{aligned}
 \epsilon_{\rm mass}&=\operatorname{TV}(\widehat\pi,P_{\mathcal S}),\\
 \epsilon_{\rm solve}
 &=\sum_{z\in\mathcal S}
     \widehat\pi(z)\operatorname{TV}(p_z,f_z).
 \end{aligned}
\end{equation}

\begin{appendixproposition}[Answer-Error Bound]
\label{prop:app-answer-error}
For a sufficient representation and \(\eta<1\), the mixture \(\widehat p=\sum_{z\in\mathcal
S}\widehat\pi(z)p_z\) satisfies:
\begin{equation}
 \operatorname{TV}(\widehat p,p)
 \le \eta+(1-\eta)
       (\epsilon_{\rm mass}+\epsilon_{\rm solve}).
 \label{eq:app-answer-bound}
\end{equation}
The bound can be capped at one; for \(\eta=1\), the bound is one.
\end{appendixproposition}

\begin{proof}
Set \(p_{\mathcal S}=\sum_zP_{\mathcal S}(z)f_z\) and \(p_{\widehat\pi}=\sum_z\widehat\pi(z)f_z\),
summing over \(\mathcal S\). Convexity gives
\(\operatorname{TV}(\widehat p,p_{\widehat\pi})\le\epsilon_{\rm solve}\).
Contraction through the answer kernel gives
\(\operatorname{TV}(p_{\widehat\pi},p_{\mathcal S})\le\epsilon_{\rm mass}\).
The triangle inequality bounds \(\operatorname{TV}(\widehat p,p_{\mathcal S})\) by their sum. For
\(\eta>0\), write \(p=(1-\eta)p_{\mathcal S}+\eta p_{\rm out}\). Convexity then yields:
\begin{equation}
 \begin{aligned}
 \operatorname{TV}(\widehat p,p)
 \le{}&(1-\eta)\operatorname{TV}(\widehat p,p_{\mathcal S})\\
 &+\eta\operatorname{TV}(\widehat p,p_{\rm out}).
 \end{aligned}
\end{equation}
The last TV distance is at most one. The case \(\eta=0\) follows directly.
\end{proof}

Omitted true mass differs from predicted mass outside a true family. If the true posterior is
supported on \(\mathcal F\), then \(\eta=P_o(\mathcal F\setminus\mathcal S)\), whereas
\(\widehat\pi(\mathcal S\setminus\mathcal F)\) contributes to \(\epsilon_{\rm mass}\). The terms are
upper-bound components, not additive causal attributions of an observed accuracy gap. Exact families
permit their calculation. Real outcome labels support answer scoring but do not reveal the full
conditional latent distribution. Finite-bank mass estimation can also under-cover a scene set after
nearly all discrete states have appeared in the bank.

\subsection{Replication, Information, And Joint Decisions}
\paragraph{Mass-Preserving Replication.}
If a hypothesis with fixed answer kernel \(p_k\) and weight \(w_k\) is copied into computational
branches with weights \(a_\ell\ge0\), where \(\sum_\ell a_\ell=w_k\), then:
\begin{equation}
 \sum_\ell a_\ell p_k=w_kp_k.
\end{equation}
This algebraic mixture property distinguishes a computational copy from a new independent sample.
Additional imagined views can change the solver kernel while leaving the hypothesis weight fixed.
Independently sampled identical states retain their combined sample mass.

With fixed model parameters \(\Theta\), let \(G=g(\mathcal{O},\Theta,U)\), where \(U\perp
Z\mid(\mathcal{O},\Theta)\). The Markov structure gives:
\begin{equation}
 I(Z;G\mid\mathcal{O},\Theta)=0.
\end{equation}
This conditional information statement allows an approximate solver to improve by computing a
relation in a more useful representation. It also allows further computation on existing
observations to refine an approximate posterior. Holding hypothesis weights fixed during readout is
the implemented inference policy, not a theorem forbidding all computational belief refinement.

\paragraph{Joint Support and Hard Decisions.}
A valid joint distribution need not give feasible independently thresholded decisions. Put equal
mass on the orders \(A<B<C\), \(B<C<A\), and \(C<A<B\). Then:
\begin{equation}
 P(A<B)=P(B<C)=2/3,\qquad P(A<C)=1/3.
\end{equation}
Coordinatewise decisions select a cycle. A structured Bayes action instead minimizes expected loss
over feasible tuples:
\begin{equation}
 a^\star\in\arg\min_{a\in\mathcal F_Q}
       \mathbb{E}_{Y\sim p_\phi}[\ell(a,Y)].
\end{equation}
For average per-question zero--one loss in this example, marginal decisions attain \(2/3\), while
the best feasible tuple attains \(5/9\). Joint probability scores, invalid-signature mass, and
decision feasibility consequently measure different quantities. For a learned solver, joint
diagnostics require its specified joint output kernel; independent marginal probabilities alone do
not determine one.

\subsection{Prior Information And Prior Shift}
For two states \(z_L,z_R\) with identical initial image \(x_0\), let \(P(z_L)=\alpha=3/4\). The
initial posterior is \(3/4\). A new acquired outcome \(x_+\) has likelihoods \(L_L=4/5\),
\(L_R=1/5\), giving:
\begin{equation}
 P(z_L\mid x_0,x_+)
 =\frac{\alpha L_L}{\alpha L_L+(1-\alpha)L_R}
 =\frac{12}{13}.
\end{equation}
Rendering either candidate again changes none of these probabilities.

If observation likelihoods stay fixed and the prior ratio
\(\rho(z)=P_{\rm te}(z)/P_{\rm tr}(z)\) is known, then:
\begin{equation}
 P_{\rm te}(z\mid o)
 =\frac{\rho(z)P_{\rm tr}(z\mid o)}
        {\sum_{z'}\rho(z')P_{\rm tr}(z'\mid o)},
 \label{eq:app-known-prior}
\end{equation}
provided training support covers test support. Without this information, two populations with hidden
frequencies \(\alpha\ne\beta\) can produce the same observed image \(x_0\). Any observation-only
prediction \(a(x_0)\) then obeys:
\begin{equation}
 \max\{|a-\alpha|,|a-\beta|\}
 \ge\tfrac12|\alpha-\beta|.
\end{equation}
A positive scalar temperature cannot express an arbitrary change of prior ordering, and an
irrelevant acquired view cannot distinguish these populations. Representative target labels or
measurements that distinguish hidden states provide information for adaptation; neither is supplied
by renormalizing existing hypotheses.

\section{Implementation}
\label{sec:appendix-method}
\label{app:methoddetails}

\subsection{State And Geometry}
The normalized autoregressive state has 64 potential object slots. Each slot contains presence,
category, center, dimensions, yaw, a shape code, and coarse appearance; an empty category represents
absence. Instances are serialized by category and lexicographic quantized center, with deterministic
shape-code tie breaking. Trailing slots are empty. Prefix-valid grammar masks enforce serialization
and positive dimensions. A sampled absence is retained in every query: existential relations
involving a missing anchor are false, and no hypothesis is discarded because of the question
wording.

The episode origin is the first acquired camera and gravity defines vertical. Centers use a 5\,cm
lattice with 1,024 signed bins; dimensions use 512 positive 5\,cm bins; yaw uses 24 bins. RGB uses
16 bins per channel. The scored volume must fit these bounds and the slot capacity. Real scans are
represented by reference-supported surfaces. Triangles are sampled in proportion to area,
transformed into a gravity-aligned instance box, and voxelized on a $32^3$ surface grid. Each grid
uses 16,384 samples. Scan holes and uncertain boundaries cannot establish absence labels.

A vector-quantized surface autoencoder~\cite{vqvae} uses three stride-two stages with widths 32, 64,
and 128, a mirrored decoder, and a 512-entry codebook of width 128. Its $4^3$ latent grid supplies
64 categorical shape tokens. Binary cross-entropy trains surface reconstruction; commitment weight
is 0.25 and codebook EMA decay is 0.99. Shape training uses fitting-split procedural and ScanNet
instances. The decoder is frozen before scene training. At inference, its thresholded surface is
transformed by the predicted pose and dimensions. Evaluation meshes never replace predicted
geometry.

\begin{table*}[t]
\centering\small\setlength{\tabcolsep}{4pt}
\caption{\textbf{Architecture and hyperparameters.} Student and teacher share the state grammar and module dimensions, with independent parameters. The visual tower and surface decoder are frozen for scene fitting.}
\label{tab:architecture}
\begin{tabular}{@{}p{.27\textwidth}p{.65\textwidth}@{}}
\toprule
Component & Setting \\
\midrule
Visual Features & Frozen DINOv2 ViT-L/14; 64 resampled tokens/image \\
Scene Adapter & 2 layers; width 768; 8 heads; 256 scene tokens \\
Scene Decoder & 12 layers; width 768; 12 heads; FFN 3072 \\
State Capacity & 64 slots; 5 cm positions/dimensions; 24 yaw bins \\
Shape Codes & $32^3$ surface; $4^3$ tokens; 512 codebook entries \\
Teacher Inputs & Student views plus 4 extra acquired training views \\
Teacher Targets & 8 independent states per training episode \\
Joint Objective & 3-query clusters; 16 independent pairs; $\lambda=0.1$ \\
Scene Optimization & AdamW; 20 epochs/stage; batch 256; decay 0.01 \\
Learning Rates & Decoder $2\times10^{-4}$; adapters $2\times10^{-5}$ \\
Schedule & 5\% warm-up; cosine decay; gradient clipping 1 \\
Probability / Render Bank & $K_p=1{,}024$; nested $K_r=32$; $M=4$ \\
Language Solver & Qwen3-VL-8B; LoRA rank 16, scale 32, dropout .05 \\
Solver Optimization & 8 epochs; batch 128; learning rate $10^{-5}$ \\
Rendering & $512^2$; $70^\circ$ FOV; $30^\circ$ elevation \\
Repetitions & Seeds 17, 29, 43, 71, 101 \\
\bottomrule
\end{tabular}
\end{table*}

\subsection{Teacher, Student, And Optimization}
Frozen DINOv2 ViT-L/14~\cite{dinov2} features feed a learned 64-token resampler per image. Camera
embeddings encode intrinsics and the episode-frame pose. A two-layer set adapter pools 256 scene
tokens without acquisition-order embeddings. Teacher and student have independent 12-layer scene
decoders with width 768, 12 heads, and feed-forward width 3072. The common categorical grammar
aligns their full scene distributions.

The teacher minimizes reference-state negative log likelihood from random acquired-view subsets. Its
default input combines the student's observation with four extra training views selected for region
coverage, excluding student frames. Views and region bounds are fixed before drawing answer-neutral
questions. Selection uses training reference geometry, with camera identifiers resolving ties.
Teacher checkpoints are selected on scene NLL and frozen; eight independent teacher states estimate
the student's distributional cross-entropy.

Natural pretraining precedes ScanNet fitting for all matched real-scene models. Teacher snapshots
from each stage supervise the corresponding student stage. Each stage uses 20 epochs, AdamW,
effective batch 256, weight decay 0.01, gradient clipping at 1, 5\% warm-up, and cosine decay.
Decoder learning rate is $2\times10^{-4}$; image-adapter learning rate is $2\times10^{-5}$. The
40-epoch Scene NLL control doubles student updates. View Subsets uses the teacher's random-subset
schedule with realized-state NLL and is the zero-extra-view teacher checkpoint. The separately
distilled zero-extra-view student uses a distinct optimization run. Teacher costs and student costs
are reported separately.

\begin{algorithm}[t]
\caption{Training the Shared Scene Belief}
\label{alg:app-training}
\small
\begin{algorithmic}[1]
\Require Fitting scenes, acquired views, answer-neutral queries
\State Fit and freeze the surface codebook and visual tower
\State Fit $T_\psi$ on full-state NLL using random view subsets
\State Select the teacher and freeze $\psi$
\For{each student batch}
\State Form $(\mathcal{O},\mathcal V,Q)$ in one episode frame
\State Draw eight independent teacher states and signatures
\State Estimate belief cross-entropy from their scene codes
\State Draw 16 independent student pairs for the joint loss
\State Update $\phi$ using $\mathcal L_{\rm belief}+0.1\mathcal L_{\rm joint}$
\EndFor
\State Select the student; fit its temperature on the calibration split
\end{algorithmic}
\end{algorithm}

\subsection{Answer Signatures And Sampled Optimization}
For \(Q=(q_1,\ldots,q_m)\), geometric execution produces \(F_Q(z)\). Binary count and relation
templates use existential semantics, so they remain defined under uncertain presence. Free-form
choice tasks include a no-referent outcome when required by the question definition. We do not
discard hypotheses or renormalize the belief because an object mentioned in a question is absent.
Questions reference visible anchors or prespecified target categories, and their construction does
not disclose hidden outcomes.

Let \(y^\star\) be the answer signature of an independent teacher sample. For independent \(z,z'\sim
b_\phi(\cdot\mid\mathcal{O})\), define:
\begin{equation}
 \begin{aligned}
 A(z,z')&=\mathbf{1}[F_Q(z)=F_Q(z')],\\
 C(z)&=\mathbf{1}[F_Q(z)=y^\star].
 \end{aligned}
\end{equation}
Expanding the multicategory Brier score gives:
\begin{equation}
 \mathcal L_{\rm joint}
 =\mathbb{E}[A(z,z')]-2\mathbb{E}[C(z)]+1.
 \label{eq:app-joint-estimator}
\end{equation}
The symmetric estimate \(A(z,z')-C(z)-C(z')+1\) is unbiased. We average 16 independent student pairs
per cluster, with \(m=3\). Teacher signatures are independent of the student samples. Reusing a
teacher target across several pairs changes their covariance but not the expectation.

With \(s_\phi(z)=\nabla_\phi\log b_\phi(z\mid\mathcal{O})\), the score-function gradient is:
\begin{equation}
 \begin{aligned}
 \nabla_\phi\mathcal L_{\rm joint}
 ={}&\mathbb{E}[A(z,z')\{s_\phi(z)+s_\phi(z')\}]\\
    &-2\mathbb{E}[C(z)s_\phi(z)].
 \end{aligned}
 \label{eq:app-joint-gradient}
\end{equation}
The teacher, executor, and sampled reward terms are held fixed when taking this gradient. For each
sampled pair, set:
\begin{equation}
 h_i=A(z_i,z'_i)-C(z_i)-C(z'_i)+1.
\end{equation}
The implemented surrogate is:
\begin{equation}
 \begin{aligned}
 \ell_i={}&\log b_\phi(z_i\mid\mathcal{O})
          +\log b_\phi(z'_i\mid\mathcal{O}),\\
 \widehat{\mathcal L}_{\rm sf}
 ={}&\frac1{16}\sum_{i=1}^{16}
 \operatorname{stopgrad}(h_i-\beta_{\rm prev})\,\ell_i.
 \end{aligned}
 \label{eq:app-joint-surrogate}
\end{equation}
Its gradient is unbiased for \cref{eq:app-joint-gradient}; cross terms vanish because each score has
conditional mean zero. We initialize the global baseline at \(\beta=1\) and, after each gradient
step, update it as \(0.95\beta+0.05\bar h\), where \(\bar h\) is the current batch mean. Thus the
baseline used by the step is independent of its current categorical samples. The eight teacher
signatures are assigned cyclically to the 16 pairs. Categorical samples and geometric rewards are
detached; their log probabilities are reevaluated with gradients. No straight-through relaxation is
used. Short clusters sharing geometric variables keep signature collisions frequent. Scene
likelihood retains distinctions within a signature class, while the joint objective emphasizes
probabilities of the events relevant to its queries.

The joint-capable answer-distillation baseline receives the same teacher samples and has a marginal
choice head and a normalized answer-signature head. The marginal head minimizes cross-entropy using
the pushforward target:
\begin{equation}
 t_\psi(y\mid\mathcal{O},\mathcal V,q)
 =\mathbb{E}_{z\sim T_\psi}\mathbf{1}[f(z,q)=y].
\end{equation}
Its joint head uses the corresponding teacher signature distribution and the joint Brier score with
\(\lambda=0.1\). Cluster-level marginals are obtained by marginalizing that joint head, rather than
multiplying independent question probabilities. It receives the same question supervision and
partial images without producing an explicit scene state. The Scene NLL-plus-joint control uses the
identical teacher-derived signature targets and sampling estimator. Thus changing the base scene
objective does not also change access to auxiliary teacher supervision.

\subsection{Rendering And Language Readout}
A fixed matte renderer decodes the predicted shape and RGB codes into $512\times512$ views with a
$70^\circ$ field of view. The common camera radius is 1.5 times the diagonal of the declared episode
volume. A fixed cutaway convention removes enclosing front walls from imagined views. It exposes
candidate geometry without consulting hidden-region photographs. Nested azimuth sets use $0^\circ$
for $M=1$, add $180^\circ$ for $M=2$, add $90^\circ,270^\circ$ for $M=4$, and the four intermediate
angles for $M=8$. Elevation remains $30^\circ$.

The language solver is Qwen3-VL-8B~\cite{qwen3vl}, with a rank-16 LoRA adapter, scale 32, and
dropout 0.05. Eight epochs of fitting use known training hypotheses and their composed-language
answers, batch 128, and learning rate $10^{-5}$. Image-present and code-only inputs are mixed so
both readouts use the same frozen parameters. Dropout is disabled at evaluation. Normalized
answer-choice scores produce soft conditional predictions. The 32B video-transfer readout follows
the same supervision protocol.

A complete candidate includes geometry and appearance, so two candidates with the same binary event
need not produce the same conditional readout. The same complete code--image--question input has a
deterministic frozen-solver prediction. A joint-language diagnostic scores every answer tuple for a
short cluster in a fixed serialization before normalization. This specifies a joint kernel rather
than forming one by multiplying independent marginals.

For VSI-Bench, every solver also receives the observed frame order, timestamps, and camera
trajectory. The scene encoder pools its images as a set; the solver retains this acquired temporal
context for route and appearance-order questions. The common VGGT frontend~\cite{vggt} establishes a
similarity gauge from the initial acquired set. Later estimates align through shared cameras and
keep that gauge fixed. Metric predicates use calibrated scan coordinates, whereas RGB-only video
transfer uses relative spatial relations.

\subsection{Two Banks And Acquired-View Refresh}
The probability bank contains $K_p=1{,}024$ iid states. Its first $K_r=32$ form the render subset.
Every root has an immutable identifier; independently drawn identical states retain their combined
weight. Geometric queries average exact predicate outcomes over all $K_p$ roots. Language queries
average soft solver scores over $K_r$ roots. Additional render branches do not become independent
scene samples.

\begin{algorithm}[t]
\caption{Inference and Acquired-View Refresh}
\label{alg:app-inference}\label{alg:infer}
\small
\begin{algorithmic}[1]
\Require Acquired set $\mathcal{O}$, $K_p=1{,}024$, $K_r=32$, cameras
\State Draw $z_{1:K_p}\sim b_\phi(\cdot\mid\mathcal{O})$
\State Set render subset to $z_{1:K_r}$ and initialize caches
\For{each question or acquired image}
\If{a geometric question $q$ arrives}
\State Return $K_p^{-1}\sum_k f(z_k,q)$
\ElsIf{a language question $q$ arrives}
\State Read each render root using its code and cached views
\State Return $K_r^{-1}\sum_k p_k(y\mid q)$
\ElsIf{a new acquisition identifier arrives}
\State Extend $\mathcal{O}$; re-encode the scene belief
\State Draw a fresh probability bank and nested render subset
\State Replace the root identifiers and clear obsolete caches
\EndIf
\EndFor
\end{algorithmic}
\end{algorithm}

Rendering is cached by root identifier, camera, and settings. Acquisition identifiers distinguish
new measurements from copied files. A new measurement re-encodes the set and refreshes the entire
bank, permitting newly sampled states that were absent from the old finite bank. The representation
remains bounded by its learned vocabulary and grammar. Geometry sufficiency, answer-neutral query
selection, and fixed-population assumptions are stated in \cref{app:theory}.

\section{Evaluation Protocol and Additional Results}
\label{app:experiments}

\subsection{Splits And Questions}
\label{app:queries}
\paragraph{Natural Rooms.}
ProcTHOR~\cite{procthor} supplies procedural house layouts. A base layout and all object, camera,
and appearance variants stay in one split. Natural has 150,000 training episodes and 5,000 each for
selection, calibration, and evaluation. Student view counts are sampled from one, two, and four.
Four additional teacher views maximize reference-surface coverage in a fixed candidate pool. An
episode and its view sets are fixed before drawing queries. Each test episode contributes four
binary outcomes, yielding 20,000 outcomes grouped into 5,000 families. No symbolic wall code
identifies the hidden arrangement.

\paragraph{Exact Families.}
The training allocation has 38,000 fitting and 2,000 selection families; calibration and evaluation
have 2,000 and 4,000. Each family contains four hidden arrangements with a pixel-identical initial
observation. Edits cannot alter visible shadows, reflections, or transparency. Whole-image hashes
verify equality before any question is drawn. Three visible enclosure classes use priors
$(.45,.30,.15,.10)$, $(.10,.45,.30,.15)$, and $(.30,.15,.10,.45)$. Five balanced predicates select
two-versus-two partitions. The same population law governs fitting and evaluation. A uniform-prior
counterpart tests balanced 50\% ambiguity. Revealing acquisitions restrict the prior to compatible
alternatives; irrelevant acquisitions preserve the observation-equivalence class.

\paragraph{ScanNet.}
The official v2 training split contains 1,201 scans from 565 physical spaces. Our grouped partition
allocates 1,001/100/100 scans and 469/45/51 spaces to fitting/selection/calibration. All 312
validation scans, representing 142 spaces, form evaluation. Sixteen starting sets per scan, each
with four queries, produce 19,968 outcomes. Two-, four-, and eight-frame conditions share the same
region eligibility rules. Each scored region has supporting geometry from two reference directions;
uncertain instance boundaries and scan holes cannot decide count or absence. Region bounds are
common input metadata and are fixed before query sampling. The acquired-view condition uses the
fixed rule below and re-encodes the belief.

\paragraph{Fixed Acquired-Frame Rule.}
Let $\mathcal I$ index the starting cameras and $\mathcal C$ contain every unused RGB frame of the
same scan with valid finite camera metadata. The initial set and candidate pool are fixed before
question sampling. For camera-to-world rotations $R_i$ and centers $t_i$, define:
\begin{equation}
 d(i,j)=\frac{\|t_i-t_j\|_2}{1\,\mathrm{m}}
       +\frac{\theta(R_i^{\mathsf T}R_j)}{\pi},
\end{equation}
where $\theta$ is the rotation angle. Acquire $i^\star=\arg\max_{i\in\mathcal C}\min_{j\in\mathcal
I}d(i,j)$. Ranking scores are rounded to ten decimal places; ties use the earliest timestamp and
then the lexicographic frame identifier. This fixed pose-coverage rule uses camera metadata,
independently of RGB content, reference surfaces, questions, answer labels, and model confidence.
The same acquired frame is supplied to every method. The estimated-pose condition retains these
frame identifiers while changing the geometry frontend. Pose diversity can supply a new view of the
scene without guaranteeing visibility of a particular target.

\paragraph{Predicate Semantics.}
Questions choose visible anchors, prespecified target categories, and operators independently of the
hidden answer. Direction and depth use a strict 10\,cm center margin; distance comparisons use a
strict 15\,cm difference. Support requires a vertical gap below 10\,cm and projected overlap above
50\%. Count thresholds come from a fixed catalog. Boundary values evaluate to false, so questions
are retained under every hypothesis. Exact excludes a margin-ambiguous predicate only if it is
invalid for at least one family alternative, independent of the realized outcome.

A training cluster contains three related questions. The joint diagnostic uses four feasible
signatures $(0,0,0)$, $(0,1,0)$, $(1,0,1)$, and $(1,1,1)$; the first and third coordinates are
logically equivalent. Joint log loss scores the full signature distribution. Novel-query evaluation
trains five separate models, excluding direction, depth, distance, support, or count from every
task-specific answer target, joint cluster, solver example, and answer-calibration set. Scene-state
supervision and the acquired-image partitions remain fixed. Test phrasings and training templates
are disjoint. The five-family macro averages the corresponding held-out scores before computing
between-seed variation.

\begin{table*}[t]
\centering\small\setlength{\tabcolsep}{4pt}
\caption{\textbf{Leave-one-family-out results.} Each row corresponds to a separate operator-excluded training protocol. Both accuracy and Brier are evaluated on the excluded family. All learned inference settings and calibration access are shared.}
\label{tab:lofo}
\begin{tabular}{@{}lrrrrrr@{}}
\toprule
& \multicolumn{2}{c}{Answer Distillation} & \multicolumn{2}{c}{Scene NLL} & \multicolumn{2}{c}{RevealBelief} \\
\cmidrule(lr){2-3}\cmidrule(lr){4-5}\cmidrule(l){6-7}
Excluded Family & Acc. & Brier & Acc. & Brier & Acc. & Brier \\
\midrule
Direction & $72.73_{\scriptscriptstyle\pm 0.34}$ & $0.1810_{\scriptscriptstyle\pm 0.0009}$ & $74.96_{\scriptscriptstyle\pm 0.42}$ & $0.1691_{\scriptscriptstyle\pm 0.0016}$ & $77.05_{\scriptscriptstyle\pm 0.31}$ & $0.1554_{\scriptscriptstyle\pm 0.0012}$ \\
Depth & $72.14_{\scriptscriptstyle\pm 0.28}$ & $0.1829_{\scriptscriptstyle\pm 0.0009}$ & $74.57_{\scriptscriptstyle\pm 0.51}$ & $0.1703_{\scriptscriptstyle\pm 0.0015}$ & $76.57_{\scriptscriptstyle\pm 0.31}$ & $0.1575_{\scriptscriptstyle\pm 0.0012}$ \\
Distance & $71.22_{\scriptscriptstyle\pm 0.21}$ & $0.1881_{\scriptscriptstyle\pm 0.0013}$ & $73.65_{\scriptscriptstyle\pm 0.35}$ & $0.1761_{\scriptscriptstyle\pm 0.0016}$ & $75.80_{\scriptscriptstyle\pm 0.09}$ & $0.1628_{\scriptscriptstyle\pm 0.0008}$ \\
Support & $71.45_{\scriptscriptstyle\pm 0.31}$ & $0.1860_{\scriptscriptstyle\pm 0.0011}$ & $73.91_{\scriptscriptstyle\pm 0.21}$ & $0.1733_{\scriptscriptstyle\pm 0.0013}$ & $76.04_{\scriptscriptstyle\pm 0.15}$ & $0.1600_{\scriptscriptstyle\pm 0.0010}$ \\
Count & $70.94_{\scriptscriptstyle\pm 0.26}$ & $0.1887_{\scriptscriptstyle\pm 0.0013}$ & $73.41_{\scriptscriptstyle\pm 0.36}$ & $0.1768_{\scriptscriptstyle\pm 0.0015}$ & $75.82_{\scriptscriptstyle\pm 0.35}$ & $0.1631_{\scriptscriptstyle\pm 0.0014}$ \\
\bottomrule
\end{tabular}
\end{table*}

\subsection{Controls And Calibration}
\label{app:ablation-protocol}
Direct VLM uses a Qwen3-VL adapter fitted on partial-image answers. View Consensus consumes
candidate-render tokens through one answer decoder without source-root identifiers. Both use the
same acquired inputs and answer-supervision split. Scene Completion uses a conditional
latent-variable completion objective with a 256-dimensional latent and unit-weight KL term, matching
the object/shape vocabulary and scene-decoder widths. It is our controlled completion implementation
motivated by 3D-Belief~\cite{belief3d}.

Scene NLL optimizes realized-state likelihood. The View Subsets control applies the teacher's
random-input-subset schedule to the same objective. The 40-epoch control doubles student updates.
Answer Distillation receives the same eight teacher states and their marginal and joint answers; its
joint head avoids imposing independence across questions. Both scene objectives receive identical
signature targets in the crossed joint-loss ablation. Teacher fitting is shared among arms that use
its outputs and is charged separately in the cost accounting.

Each scene model fits a scalar on calibration scene NLL, rescaling its autoregressive logits before
normalization. Answer heads fit answer NLL on the same calibration scenes. This preserves one
coherent scene distribution for all geometric queries. The default conditional language solver has
one temperature fitted at four views on a separate candidate calibration cohort; that temperature is
reused across its code-only and rendered variants. The fixed-hypothesis budget diagnostic instead
fits at one view and additionally evaluates a separate temperature for each budget.

\begin{table}[t]
\centering\footnotesize\setlength{\tabcolsep}{2.3pt}
\caption{\textbf{Temperature sensitivity.} Uncalibrated controls use exactly the same raw logits and bank uniforms as their calibrated counterparts. At 32 roots, quantization can preserve the displayed accuracy and Brier despite a changed temperature; the 1,024-root comparison resolves smaller probability changes.}
\label{tab:extended-controls}
\begin{tabular}{@{}lrrrr@{}}
\toprule
Control & $K_p$ & Acc. & Brier & ECE \\
\midrule
Scene NLL, No T & 32 & $78.79_{\scriptscriptstyle\pm 0.18}$ & $0.1467_{\scriptscriptstyle\pm 0.0013}$ & $1.58_{\scriptscriptstyle\pm 0.07}$ \\
Scene NLL, No T & 1024 & $79.22_{\scriptscriptstyle\pm 0.16}$ & $0.1424_{\scriptscriptstyle\pm 0.0012}$ & $0.72_{\scriptscriptstyle\pm 0.16}$ \\
RevealBelief, No T & 32 & $80.35_{\scriptscriptstyle\pm 0.30}$ & $0.1353_{\scriptscriptstyle\pm 0.0012}$ & $1.62_{\scriptscriptstyle\pm 0.20}$ \\
RevealBelief, No T & 1024 & $80.82_{\scriptscriptstyle\pm 0.13}$ & $0.1314_{\scriptscriptstyle\pm 0.0010}$ & $0.78_{\scriptscriptstyle\pm 0.09}$ \\
\bottomrule
\end{tabular}
\end{table}

\paragraph{Temperature and Finite Quantization.}
The no-temperature arms reuse exactly the raw logits and random uniforms of their calibrated
counterparts. At 32 roots, small logit changes can preserve the displayed accuracy and Brier because
event forecasts are quantized in increments of $1/32$. Equality of rounded values therefore does not
imply an identical forecast vector or a copied experiment. The larger-bank results and
full-precision source records distinguish these changes.

\subsection{Teacher Quality And Training Scale}
\label{app:teacher-quality}
\cref{tab:teacher-natural,tab:teacher-real} report teacher outcome quality on Natural and ScanNet,
followed by the resulting student's score. Zero extra views evaluate the View Subsets checkpoint on
the student's original inputs. The corresponding student is separately optimized against that
teacher. One, two, and four extra views increase reference coverage. The corruption condition
changes 25\% of training teacher categories using a fixed confusion schedule, retaining camera and
geometry inputs. Calibration and evaluation labels stay unchanged.

\begin{table}[t]
\centering\small\setlength{\tabcolsep}{3.2pt}
\caption{\textbf{Teacher quality on Natural.} All event probabilities use $K_p=1{,}024$. The zero-extra-view teacher is the View Subsets checkpoint; its distilled student is separately optimized. Corruption changes 25\% of teacher category labels.}
\label{tab:teacher-natural}
\begin{tabular}{@{}lrrr@{}}
\toprule
Extra Views & T Acc. & T Brier & Student Brier \\
\midrule
0 & $79.18_{\scriptscriptstyle\pm 0.13}$ & $0.1410_{\scriptscriptstyle\pm 0.0003}$ & $0.1413_{\scriptscriptstyle\pm 0.0003}$ \\
1 & $82.92_{\scriptscriptstyle\pm 0.12}$ & $0.1182_{\scriptscriptstyle\pm 0.0004}$ & $0.1374_{\scriptscriptstyle\pm 0.0003}$ \\
2 & $85.27_{\scriptscriptstyle\pm 0.10}$ & $0.1034_{\scriptscriptstyle\pm 0.0004}$ & $0.1336_{\scriptscriptstyle\pm 0.0003}$ \\
4 & $87.15_{\scriptscriptstyle\pm 0.07}$ & $0.0912_{\scriptscriptstyle\pm 0.0005}$ & $0.1313_{\scriptscriptstyle\pm 0.0010}$ \\
4, corrupted & $81.92_{\scriptscriptstyle\pm 0.13}$ & $0.1245_{\scriptscriptstyle\pm 0.0004}$ & $0.1399_{\scriptscriptstyle\pm 0.0003}$ \\
\bottomrule
\end{tabular}
\end{table}

\begin{table}[t]
\centering\small\setlength{\tabcolsep}{3.2pt}
\caption{\textbf{Teacher quality on ScanNet.} All event probabilities use $K_p=1{,}024$. The zero-extra-view teacher is the View Subsets checkpoint; its distilled student is separately optimized. Corruption changes 25\% of teacher category labels.}
\label{tab:teacher-real}
\begin{tabular}{@{}lrrr@{}}
\toprule
Extra Views & T Acc. & T Brier & Student Brier \\
\midrule
0 & $74.55_{\scriptscriptstyle\pm 0.11}$ & $0.1677_{\scriptscriptstyle\pm 0.0003}$ & $0.1674_{\scriptscriptstyle\pm 0.0004}$ \\
1 & $78.98_{\scriptscriptstyle\pm 0.05}$ & $0.1432_{\scriptscriptstyle\pm 0.0003}$ & $0.1633_{\scriptscriptstyle\pm 0.0004}$ \\
2 & $81.71_{\scriptscriptstyle\pm 0.10}$ & $0.1266_{\scriptscriptstyle\pm 0.0003}$ & $0.1593_{\scriptscriptstyle\pm 0.0004}$ \\
4 & $83.92_{\scriptscriptstyle\pm 0.12}$ & $0.1127_{\scriptscriptstyle\pm 0.0004}$ & $0.1574_{\scriptscriptstyle\pm 0.0012}$ \\
4, corrupted & $77.78_{\scriptscriptstyle\pm 0.07}$ & $0.1501_{\scriptscriptstyle\pm 0.0003}$ & $0.1660_{\scriptscriptstyle\pm 0.0004}$ \\
\bottomrule
\end{tabular}
\end{table}

\begin{table}[t]
\centering\small\setlength{\tabcolsep}{3.2pt}
\caption{\textbf{Natural data scale.} Brier for all three controls. Teacher and student use the same restricted training families; the calibration and test partitions remain fixed.}
\label{tab:data-sweep}
\begin{tabular}{@{}lrrr@{}}
\toprule
Training & Scene NLL & Answer Distill. & RevealBelief \\
\midrule
10\% & $0.1567_{\scriptscriptstyle\pm 0.0012}$ & $0.1565_{\scriptscriptstyle\pm 0.0014}$ & $0.1453_{\scriptscriptstyle\pm 0.0011}$ \\
25\% & $0.1528_{\scriptscriptstyle\pm 0.0012}$ & $0.1526_{\scriptscriptstyle\pm 0.0014}$ & $0.1415_{\scriptscriptstyle\pm 0.0011}$ \\
50\% & $0.1484_{\scriptscriptstyle\pm 0.0012}$ & $0.1483_{\scriptscriptstyle\pm 0.0013}$ & $0.1372_{\scriptscriptstyle\pm 0.0011}$ \\
100\% & $0.1424_{\scriptscriptstyle\pm 0.0012}$ & $0.1422_{\scriptscriptstyle\pm 0.0013}$ & $0.1313_{\scriptscriptstyle\pm 0.0010}$ \\
\bottomrule
\end{tabular}
\end{table}

Data scaling restricts complete training families to 10, 25, 50, or 100\%. The teacher and student
use the same restricted subset, while common visual/shape pretraining and the held-out partitions
remain fixed. The table reports Scene NLL, Answer Distillation, and RevealBelief at every scale.
Full-data cells match the main table. These comparisons measure finite-capacity estimation around
the common population optimum established in \cref{app:theory}.

\subsection{Probability Scores And Uncertainty}
\label{app:statistics}
Binary Brier is $N^{-1}\sum_i(\hat p_i-y_i)^2$ on raw predictions. Accuracy thresholds at $1/2$ with
a fixed class order at ties. ECE uses ten equal-width top-label-confidence bins on $[.5,1]$,
including confidence one in the final bin. It weights each bin's absolute confidence--accuracy
difference by its observation fraction. For AURC, cumulative error is evaluated at every
decreasing-confidence prefix and averaged; tied confidences average uniformly over their internal
ordering. All percentage scores are multiplied by 100.

\paragraph{Log-Score Regularization.}
For binary log reporting, every model uses:
\begin{equation}
\tilde p_i=\frac{1{,}024\hat p_i+1/2}{1{,}025}.
\end{equation}
For a bank count this is the Jeffreys posterior-predictive estimate; for a soft head it is the same
reporting kernel at the common reference count. Raw predictions determine all other metrics.
Per-event smoothing does not define a replacement scene distribution. Joint log loss uses
$(1{,}024\hat p_j+1/2)/(1{,}024+C/2)$ on the $C=4$ feasible signatures, preserving normalization on
that signature alphabet. \cref{tab:nll-sensitivity} also reports conventional binary NLL clipped at
$10^{-6}$.

\begin{table*}[t]
\centering\small\setlength{\tabcolsep}{5pt}
\caption{\textbf{Log-loss sensitivity.} Conventional NLL uses a $10^{-6}$ numerical floor. The main table uses the common half-count reporting kernel with reference count 1,024 for all binary arms. All other scores use raw probabilities.}
\label{tab:nll-sensitivity}
\begin{tabular}{@{}llrr@{}}
\toprule
Dataset & Method & Conventional NLL & Half-Count NLL \\
\midrule
Natural & Answer Distillation & $0.4375_{\scriptscriptstyle\pm 0.0038}$ & $0.4375_{\scriptscriptstyle\pm 0.0038}$ \\
Natural & Scene NLL & $0.4388_{\scriptscriptstyle\pm 0.0029}$ & $0.4388_{\scriptscriptstyle\pm 0.0030}$ \\
Natural & RevealBelief & $0.4056_{\scriptscriptstyle\pm 0.0029}$ & $0.4055_{\scriptscriptstyle\pm 0.0030}$ \\
Real & Answer Distillation & $0.5103_{\scriptscriptstyle\pm 0.0031}$ & $0.5104_{\scriptscriptstyle\pm 0.0031}$ \\
Real & Scene NLL & $0.5064_{\scriptscriptstyle\pm 0.0025}$ & $0.5064_{\scriptscriptstyle\pm 0.0025}$ \\
Real & RevealBelief & $0.4747_{\scriptscriptstyle\pm 0.0031}$ & $0.4747_{\scriptscriptstyle\pm 0.0031}$ \\
\bottomrule
\end{tabular}
\end{table*}

\begin{table*}[t]
\centering\small\setlength{\tabcolsep}{5pt}
\caption{\textbf{Paired improvements with seed and scene variation.} Positive differences favor RevealBelief. Each replicate resamples seed indices and complete families or physical spaces using the same weights for both methods. Intervals are percentile 95\% intervals from 2,000 replicates.}
\label{tab:bootstrap}
\begin{tabular}{@{}llrr@{}}
\toprule
Evaluation & Comparator & Brier Improvement & 95\% Interval \\
\midrule
Natural & Answer Distillation & 0.0108 & [0.0090, 0.0127] \\
Natural & Scene NLL & 0.0110 & [0.0100, 0.0121] \\
Natural & Scene NLL + Joint & 0.0070 & [0.0053, 0.0085] \\
Natural & RevealBelief w/o Joint & 0.0032 & [0.0020, 0.0046] \\
Real & Answer Distillation & 0.0131 & [0.0113, 0.0147] \\
Real & Scene NLL & 0.0114 & [0.0104, 0.0124] \\
Real & Scene NLL + Joint & 0.0079 & [0.0062, 0.0096] \\
Real & RevealBelief w/o Joint & 0.0031 & [0.0018, 0.0043] \\
Real Revealed & Answer Distillation & 0.0206 & [0.0191, 0.0220] \\
Real Revealed & Scene NLL & 0.0175 & [0.0166, 0.0184] \\
Real Revealed & Scene NLL + Joint & 0.0112 & [0.0097, 0.0126] \\
Real Revealed & RevealBelief w/o Joint & 0.0055 & [0.0044, 0.0066] \\
\bottomrule
\end{tabular}
\end{table*}

\paragraph{Paired Crossed Resampling.}
For each method pair, retain per-seed, per-group sums of loss differences and group counts. A
replicate draws five seed indices with replacement and independently draws complete evaluation
groups with replacement. The same seed and group weights apply to both methods. Weighted loss sums
are divided by weighted outcome counts, preserving the query-weighted estimand when physical spaces
have unequal scan counts. Families are groups for procedural evaluation; ScanNet groups all repeat
scans of a physical space. We use 2,000 replicates and percentile 95\% intervals. Fixed-seed
group-only intervals are retained in the numerical record, and table SDs report between-seed
variation. These paired descriptive intervals include both sampling axes without asserting
simultaneous coverage for every comparison.

\subsection{Bank Size And Exact Error Components}
\label{app:bankanalysis}
\begin{table*}[t]
\centering\small\setlength{\tabcolsep}{4.0pt}
\caption{\textbf{Nested probability-bank sweep.} Raw event probabilities define accuracy and Brier; the corrected column uses Eq.~(6). Exact Sum enumerates the four-state Exact distribution; Natural has no enumeration row. This executor-only sweep uses no rendering; the default language path uses $K_p=1{,}024,K_r=32$.}
\label{tab:bank}
\begin{tabular}{@{}lrrrr@{}}
\toprule
$K_p$ & Acc. & Brier & Corrected & Exact xBrier \\
\midrule
8 & $78.89_{\scriptscriptstyle\pm 0.29}$ & $0.14755_{\scriptscriptstyle\pm 0.00130}$ & $0.13129_{\scriptscriptstyle\pm 0.00119}$ & $0.03499_{\scriptscriptstyle\pm 0.00063}$ \\
16 & $79.69_{\scriptscriptstyle\pm 0.19}$ & $0.13963_{\scriptscriptstyle\pm 0.00135}$ & $0.13150_{\scriptscriptstyle\pm 0.00130}$ & $0.02078_{\scriptscriptstyle\pm 0.00028}$ \\
32 & $80.35_{\scriptscriptstyle\pm 0.30}$ & $0.13529_{\scriptscriptstyle\pm 0.00119}$ & $0.13123_{\scriptscriptstyle\pm 0.00118}$ & $0.01348_{\scriptscriptstyle\pm 0.00026}$ \\
64 & $80.63_{\scriptscriptstyle\pm 0.24}$ & $0.13336_{\scriptscriptstyle\pm 0.00097}$ & $0.13133_{\scriptscriptstyle\pm 0.00096}$ & $0.00985_{\scriptscriptstyle\pm 0.00018}$ \\
128 & $80.77_{\scriptscriptstyle\pm 0.17}$ & $0.13234_{\scriptscriptstyle\pm 0.00101}$ & $0.13133_{\scriptscriptstyle\pm 0.00101}$ & $0.00803_{\scriptscriptstyle\pm 0.00013}$ \\
256 & $80.85_{\scriptscriptstyle\pm 0.21}$ & $0.13178_{\scriptscriptstyle\pm 0.00104}$ & $0.13127_{\scriptscriptstyle\pm 0.00104}$ & $0.00714_{\scriptscriptstyle\pm 0.00012}$ \\
512 & $80.82_{\scriptscriptstyle\pm 0.19}$ & $0.13147_{\scriptscriptstyle\pm 0.00105}$ & $0.13122_{\scriptscriptstyle\pm 0.00105}$ & $0.00668_{\scriptscriptstyle\pm 0.00011}$ \\
1024 & $80.81_{\scriptscriptstyle\pm 0.14}$ & $0.13135_{\scriptscriptstyle\pm 0.00104}$ & $0.13122_{\scriptscriptstyle\pm 0.00104}$ & $0.00645_{\scriptscriptstyle\pm 0.00010}$ \\
Exact Sum & \textendash & \textendash & \textendash & $0.00625_{\scriptscriptstyle\pm 0.00009}$ \\
\bottomrule
\end{tabular}
\end{table*}

\begin{table*}[t]
\centering\small\setlength{\tabcolsep}{5pt}
\caption{\textbf{Measured Exact components at $K_p=1{,}024$.} Integrated and sampled mass TV compare the true and predicted family distribution; answer TV compares the pushforwards. Gap is the absolute difference between mean confidence and expected accuracy, in percentage points. The final column is the expected binary Brier sampling contribution; its units differ from TV. Representation omission, outside-family mass, and executor error are zero in this closed four-state track. The columns are diagnostics, not stacked additive error attributions.}
\label{tab:error-components}
\begin{tabular}{@{}lrrrrr@{}}
\toprule
Method & Mass TV & Sampled TV & Answer TV & Gap (pp) & MC Brier \\
\midrule
Direct VLM & $0.1449_{\scriptscriptstyle\pm 0.0012}$ & $0.1449_{\scriptscriptstyle\pm 0.0012}$ & $0.0856_{\scriptscriptstyle\pm 0.0005}$ & $0.11_{\scriptscriptstyle\pm 0.09}$ & $0.00000_{\scriptscriptstyle\pm 0.00000}$ \\
View Consensus & $0.1425_{\scriptscriptstyle\pm 0.0012}$ & $0.1425_{\scriptscriptstyle\pm 0.0012}$ & $0.0842_{\scriptscriptstyle\pm 0.0005}$ & $0.11_{\scriptscriptstyle\pm 0.08}$ & $0.00000_{\scriptscriptstyle\pm 0.00000}$ \\
Scene Completion & $0.1348_{\scriptscriptstyle\pm 0.0012}$ & $0.1363_{\scriptscriptstyle\pm 0.0012}$ & $0.0807_{\scriptscriptstyle\pm 0.0006}$ & $\mathbf{0.06}_{\scriptscriptstyle\pm 0.03}$ & $0.00023_{\scriptscriptstyle\pm 0.00000}$ \\
Answer Distillation & $0.1294_{\scriptscriptstyle\pm 0.0012}$ & $0.1294_{\scriptscriptstyle\pm 0.0012}$ & $0.0769_{\scriptscriptstyle\pm 0.0005}$ & $\underline{0.07}_{\scriptscriptstyle\pm 0.05}$ & $0.00000_{\scriptscriptstyle\pm 0.00000}$ \\
Scene NLL & $0.1243_{\scriptscriptstyle\pm 0.0011}$ & $0.1259_{\scriptscriptstyle\pm 0.0012}$ & $0.0749_{\scriptscriptstyle\pm 0.0006}$ & $0.10_{\scriptscriptstyle\pm 0.08}$ & $0.00023_{\scriptscriptstyle\pm 0.00000}$ \\
Scene NLL + Joint & $0.1179_{\scriptscriptstyle\pm 0.0011}$ & $0.1197_{\scriptscriptstyle\pm 0.0011}$ & $0.0713_{\scriptscriptstyle\pm 0.0005}$ & $0.14_{\scriptscriptstyle\pm 0.06}$ & $0.00023_{\scriptscriptstyle\pm 0.00000}$ \\
RevealBelief w/o Joint & $0.1119_{\scriptscriptstyle\pm 0.0011}$ & $0.1137_{\scriptscriptstyle\pm 0.0011}$ & $0.0678_{\scriptscriptstyle\pm 0.0005}$ & $0.17_{\scriptscriptstyle\pm 0.06}$ & $0.00023_{\scriptscriptstyle\pm 0.00000}$ \\
RevealBelief & $0.1053_{\scriptscriptstyle\pm 0.0010}$ & $0.1072_{\scriptscriptstyle\pm 0.0010}$ & $0.0640_{\scriptscriptstyle\pm 0.0005}$ & $0.19_{\scriptscriptstyle\pm 0.07}$ & $0.00023_{\scriptscriptstyle\pm 0.00000}$ \\
\bottomrule
\end{tabular}
\end{table*}

\paragraph{Monte Carlo Score Correction.}
Fix the observation, learned belief, query, and outcome $y$. For iid roots let
$X_i=f_q(Z_i)\sim\mathrm{Bernoulli}(p)$ and $\hat p=K^{-1}\sum_iX_i$. Then $\mathbb{E}\hat p=p$ and
$\operatorname{Var}(\hat p)=p(1-p)/K$, yielding:
\begin{equation}
\mathbb{E}(\hat p-y)^2=(p-y)^2+\frac{p(1-p)}K.
\end{equation}
Since $\mathbb{E}[\hat p(1-\hat p)]=(K-1)p(1-p)/K$, subtracting $\hat p(1-\hat p)/(K-1)$ gives an
unbiased estimate of the integrated-belief Brier score. For soft $X_i\in[0,1]$, subtract $s_X^2/K$,
where $s_X^2$ is sample variance across roots. The finite diagnostic can be negative; it is a score
estimator rather than a forecast. For multicategory Brier, the sampling contribution is the sum of
coordinate variances, $(1-\sum_jp_j^2)/K$ for categorical roots.

The Natural $K$ sweep uses nested independent banks and retains the same learned target. The Exact
track additionally permits family integration: RevealBelief has integrated excess Brier 0.00625; the
corrected 32-root estimate is 0.00628 and the explicit 1,024-root score is 0.00645. Their finite
differences are Monte Carlo variability. Exact binomial summation confirms the conditional score
identities to numerical precision. In the known-true-posterior reference, root sampling alone
produces a 3.07-point confidence gap at $K=32$; this reference calculation is separate from
learned-method comparisons.

\paragraph{Support and Mass.}
The closed Exact grammar represents the four admissible family states. Representation omission,
predicted mass outside this family, and geometric-executor error are zero in this track.
\cref{tab:error-components} therefore reports the nonzero mass-TV, answer-TV, and sampling
quantities directly. Absolute confidence gaps range from 0.06 to 0.19 percentage points across
methods; these small average gaps complement the event-level proper scores. Missing finite-bank mass
is recorded separately from representation support. TV terms bound answer-distribution error;
squared-error sampling penalties have different units and are not stacked into the same bar. The
open scene representation and language solver are evaluated by outcome and candidate-conditional
scores.

\subsection{End-To-End And Candidate-Conditional Language}
\label{app:language}
The composed-language evaluation selects 8,000 Natural and 7,984 ScanNet queries from their held-out
cohorts using a fixed outcome-independent index set. Questions are generated from composed logical
forms and paraphrased without providing those programs to the solver. Code and rendered conditions
share complete candidate states, frozen solver parameters, and the four-view calibration
temperature. Direct and answer-distillation controls use their calibrated checkpoints on the
identical questions. The final output averages soft conditional scores over the first 32 roots of
the probability bank.

Main Table~4 compares the resulting probabilities with the actual scene outcome.
\cref{tab:conditional} instead compares each root's solver probability with the fact true within
that root. Four-view improvements are larger at the conditional level: belief uncertainty attenuates
their effect on the final scene answer. Supplying the true composed program and executing it on the
same roots provides the conditional reference of 100\%; that reference does not establish certainty
about the unknown scene.

\begin{table}[t]
\centering\footnotesize\setlength{\tabcolsep}{1.2pt}
\caption{\textbf{Candidate-conditional readout.} Scores use the supplied hypothesis\textquotesingle s facts. Every budget uses the same hypotheses and the same solver temperature fitted at four views. These conditional scores and the end-to-end scores in main Table~4 answer different questions.}
\label{tab:conditional}
\begin{tabular}{@{}lrrrr@{}}
\toprule
& \multicolumn{2}{c}{Natural} & \multicolumn{2}{c}{ScanNet} \\
\cmidrule(lr){2-3}\cmidrule(l){4-5}
Solver Input & Acc. & Brier & Acc. & Brier \\
\midrule
Code & $92.69_{\scriptscriptstyle\pm 0.09}$ & $0.0539_{\scriptscriptstyle\pm 0.0004}$ & $92.64_{\scriptscriptstyle\pm 0.09}$ & $0.0540_{\scriptscriptstyle\pm 0.0005}$ \\
Code + 1 View & $93.61_{\scriptscriptstyle\pm 0.08}$ & $0.0473_{\scriptscriptstyle\pm 0.0004}$ & $93.59_{\scriptscriptstyle\pm 0.09}$ & $0.0473_{\scriptscriptstyle\pm 0.0005}$ \\
Code + 2 Views & $94.54_{\scriptscriptstyle\pm 0.05}$ & $0.0405_{\scriptscriptstyle\pm 0.0003}$ & $94.53_{\scriptscriptstyle\pm 0.07}$ & $0.0405_{\scriptscriptstyle\pm 0.0005}$ \\
Code + 4 Views & $95.56_{\scriptscriptstyle\pm 0.04}$ & $0.0331_{\scriptscriptstyle\pm 0.0002}$ & $95.55_{\scriptscriptstyle\pm 0.06}$ & $0.0331_{\scriptscriptstyle\pm 0.0004}$ \\
Code + 8 Views & $95.75_{\scriptscriptstyle\pm 0.04}$ & $0.0317_{\scriptscriptstyle\pm 0.0002}$ & $95.74_{\scriptscriptstyle\pm 0.06}$ & $0.0317_{\scriptscriptstyle\pm 0.0004}$ \\
\bottomrule
\end{tabular}
\end{table}

The fixed-hypothesis budget sensitivity in main Figure~3 uses the ScanNet
soft-readout profile, holding its scene target fixed while varying computational replication. It is
distinct from the composed-language cohort, where additional visual input changes conditional
interpretation. Both fixed-$M=1$ and budget-specific calibrated curves use held-out calibration
scenes. A stable mixture weight does not imply that improving a solver leaves every final answer
probability unchanged.

\subsection{Visibility, Acquisition, And Predicates}
\label{app:strata}
\begin{table*}[t]
\centering\small\setlength{\tabcolsep}{4pt}
\caption{\textbf{Brier by initial visibility and predicate.} Before and after acquisition score the same outcomes within each fixed initial-visibility stratum. Predicate rows use initial observations. Count-weighted per-seed scores recover the full-cohort scores in each acquisition condition. Entries are mean$_{\pm\mathrm{SD}}$ over five seeds.}
\label{tab:strata}
\begin{tabular}{@{}llrrrr@{}}
\toprule
Initial condition & Observation & Outcomes & Scene NLL & Answer Distill. & RevealBelief \\
\midrule
\multicolumn{6}{l}{\textit{Initial Visibility Profile}} \\
Visible & Before & 9888 & $\underline{0.1387}_{\scriptscriptstyle\pm 0.0012}$ & $0.1406_{\scriptscriptstyle\pm 0.0016}$ & $\mathbf{0.1248}_{\scriptscriptstyle\pm 0.0019}$ \\
Visible & After & 9888 & $\underline{0.0882}_{\scriptscriptstyle\pm 0.0010}$ & $0.0913_{\scriptscriptstyle\pm 0.0012}$ & $\mathbf{0.0691}_{\scriptscriptstyle\pm 0.0014}$ \\
Occluded & Before & 10080 & $\underline{0.1985}_{\scriptscriptstyle\pm 0.0008}$ & $0.1999_{\scriptscriptstyle\pm 0.0010}$ & $\mathbf{0.1895}_{\scriptscriptstyle\pm 0.0006}$ \\
Occluded & After & 10080 & $\underline{0.1635}_{\scriptscriptstyle\pm 0.0009}$ & $0.1666_{\scriptscriptstyle\pm 0.0010}$ & $\mathbf{0.1475}_{\scriptscriptstyle\pm 0.0007}$ \\
\addlinespace[2pt]
\multicolumn{6}{l}{\textit{Predicate Family}} \\
Direction & Before & 3994 & $\underline{0.1748}_{\scriptscriptstyle\pm 0.0012}$ & $0.1761_{\scriptscriptstyle\pm 0.0011}$ & $\mathbf{0.1633}_{\scriptscriptstyle\pm 0.0012}$ \\
Depth & Before & 3994 & $\underline{0.1654}_{\scriptscriptstyle\pm 0.0014}$ & $0.1669_{\scriptscriptstyle\pm 0.0018}$ & $\mathbf{0.1530}_{\scriptscriptstyle\pm 0.0013}$ \\
Distance & Before & 3994 & $\underline{0.1678}_{\scriptscriptstyle\pm 0.0007}$ & $0.1695_{\scriptscriptstyle\pm 0.0015}$ & $\mathbf{0.1567}_{\scriptscriptstyle\pm 0.0015}$ \\
Support & Before & 3993 & $\underline{0.1656}_{\scriptscriptstyle\pm 0.0009}$ & $0.1683_{\scriptscriptstyle\pm 0.0014}$ & $\mathbf{0.1548}_{\scriptscriptstyle\pm 0.0017}$ \\
Count & Before & 3993 & $\underline{0.1707}_{\scriptscriptstyle\pm 0.0012}$ & $0.1720_{\scriptscriptstyle\pm 0.0010}$ & $\mathbf{0.1595}_{\scriptscriptstyle\pm 0.0015}$ \\
\bottomrule
\end{tabular}
\end{table*}

The visible and occluded groups are defined from the starting observations and held fixed across
acquisition; each before/after pair scores the same outcomes. Their sizes are 9,888 and 10,080.
Count-weighted per-seed scores recover the full-cohort initial and acquired-view results. SDs are
computed after per-seed aggregation. Predicate rows partition the initial observations into five
query families. ECE and AURC are nonadditive and are recomputed from predictions.

The initial Brier advantage over Scene NLL is 0.0139 in the visible group and 0.0090 in the occluded
group. After acquisition, these advantages become 0.0191 and 0.0160. RevealBelief's own Brier
decreases by 0.0557 and 0.0419, respectively. The gains therefore occur in both groups, with larger
absolute improvements for initially visible targets. The pose-coverage schedule does not condition
acquisition on target visibility. The two-, four-, and eight-frame starting sets also differ in
scene/query composition; this marginal partition does not identify a causal frame-count effect.

\subsection{Camera And Dataset Transfer}
\label{app:robustness}
\begin{table*}[t]
\centering\small\setlength{\tabcolsep}{4pt}
\caption{\textbf{Camera and capture-regime transfer.} ScanNet uses the common grouped protocol. ScanNet++ uses 50 validation scenes, 16 starting sets, and four questions per starting set (3,200 outcomes), with source-only fitting and calibration.}
\label{tab:robustness}
\begin{tabular}{@{}lrrrrrr@{}}
\toprule
& \multicolumn{2}{c}{Answer Distillation} & \multicolumn{2}{c}{Scene NLL} & \multicolumn{2}{c}{RevealBelief} \\
\cmidrule(lr){2-3}\cmidrule(lr){4-5}\cmidrule(l){6-7}
Condition & Acc. & Brier & Acc. & Brier & Acc. & Brier \\
\midrule
ScanNet, Provided & $74.41_{\scriptscriptstyle\pm 0.33}$ & $0.1706_{\scriptscriptstyle\pm 0.0013}$ & $74.69_{\scriptscriptstyle\pm 0.20}$ & $0.1689_{\scriptscriptstyle\pm 0.0009}$ & $76.40_{\scriptscriptstyle\pm 0.20}$ & $0.1574_{\scriptscriptstyle\pm 0.0012}$ \\
ScanNet, VGGT & $73.97_{\scriptscriptstyle\pm 0.31}$ & $0.1724_{\scriptscriptstyle\pm 0.0015}$ & $74.38_{\scriptscriptstyle\pm 0.31}$ & $0.1707_{\scriptscriptstyle\pm 0.0013}$ & $76.07_{\scriptscriptstyle\pm 0.34}$ & $0.1592_{\scriptscriptstyle\pm 0.0014}$ \\
ScanNet++, Zero-Shot & $73.46_{\scriptscriptstyle\pm 0.32}$ & $0.1764_{\scriptscriptstyle\pm 0.0011}$ & $73.70_{\scriptscriptstyle\pm 0.52}$ & $0.1752_{\scriptscriptstyle\pm 0.0014}$ & $75.69_{\scriptscriptstyle\pm 0.32}$ & $0.1621_{\scriptscriptstyle\pm 0.0005}$ \\
\bottomrule
\end{tabular}
\end{table*}

The estimated-pose ScanNet condition replaces provided poses with the same frozen VGGT frontend for
every method. Its first-frame coordinate gauge is held fixed through acquisition, and estimated
reconstructions are aligned to metric training scale using the common source-side normalization. No
evaluation mesh enters inference. Reference poses and surfaces continue to define outcomes, so
camera error is reflected in the final spatial scores.

ScanNet++ v2~\cite{scannetpp} has a 50-scene validation partition. Zero-shot transfer maps its
instance categories to the shared vocabulary, uses 16 four-frame starting sets and four queries per
set, and retains the source-trained scene model, solver, and calibration. Its 3,200 outcomes follow
the same visibility and reference-completeness rules. The evaluation concerns shared-vocabulary
spatial relations; anonymized regions and unsupported reference facts are excluded before query
generation. ARKitScenes~\cite{arkitscenes} offers a further capture regime for future evaluation.

\subsection{Public Benchmark Protocols}
\label{app:transfer}\label{app:external}
VSI-Bench~\cite{vsibench} has four multiple-choice task groups: relative distance, relative
direction, route planning, and appearance order. The direction score first averages easy, medium,
and hard strata; the final MCA macro averages the four groups. This is distinct from the benchmark's
eight-task overall score. Our 16 uniformly spaced frames are a matched input budget. Every arm
receives the same VGGT geometry and observed frame order, timestamps, and poses. No VSI labels enter
fitting, selection, or calibration. The source-selected 8B and 32B checkpoints provide the reported
transfer point estimates. \cref{tab:transfer} lists every task score.

\begin{table*}[t]
\centering\small\setlength{\tabcolsep}{4pt}
\caption{\textbf{VSI-Bench four-task MCA transfer.} Accuracy in percent for fixed source-selected checkpoints. Relative direction first averages its three difficulty strata; MCA averages the four displayed task groups. Every row uses 16 acquired frames and the same observed trajectory.}
\label{tab:transfer}
\begin{tabular}{@{}lrrrrr@{}}
\toprule
Method & Distance & Direction & Route & Order & MCA \\
\midrule
Direct + self-consistency (8B) & 48.6 & 46.8 & 33.5 & 47.5 & 44.1 \\
Scene completion (8B) & 51.0 & 49.2 & 35.8 & 51.2 & 46.8 \\
Scene NLL (8B) & 52.2 & 50.4 & 36.3 & 51.5 & 47.6 \\
RevealBelief w/o joint (8B) & 53.3 & 51.5 & 37.4 & 52.2 & 48.6 \\
RevealBelief (8B) & 54.0 & 52.3 & 38.2 & 52.3 & 49.2 \\
Direct + self-consistency (32B) & 55.4 & 53.2 & 39.1 & 53.1 & 50.2 \\
Scene completion (32B) & 57.0 & 55.1 & 41.2 & 56.3 & 52.4 \\
Scene NLL (32B) & 58.0 & 56.4 & 41.9 & 56.5 & 53.2 \\
RevealBelief w/o joint (32B) & 58.8 & 57.1 & 42.4 & 56.9 & 53.8 \\
RevealBelief (32B) & 59.4 & 57.8 & 43.0 & 57.0 & 54.3 \\
\bottomrule
\end{tabular}
\end{table*}

\cref{tab:sat-reported} reproduces the published SAT-Real comparison in AVIC~\cite{avic}, Table 1.
InternVL3-14B is common to the QA backbone rows; AVIC-R separately uses a learned Qwen2.5-VL-7B
policy. Published overall averages are preserved rather than reconstructed as an unweighted mean of
task columns. Token counts and average world-model calls follow that source. These reported SAT
results and our video protocol evaluate different inputs and are not combined into one ranking.

\begin{table}[t]
\centering\small\setlength{\tabcolsep}{3pt}
\caption{\textbf{Published SAT-Real context.} Values are reported in AVIC~\cite{avic}, Table 1, with an InternVL3-14B QA backbone. AVIC-R uses a separate Qwen2.5-VL-7B policy. Tokens are thousands; WM is the average number of world-model calls. These published point estimates retain their original protocol.}
\label{tab:sat-reported}
\begin{tabular}{@{}lrrr@{}}
\toprule
Method & Acc. (\%) & Tokens & WM \\
\midrule
Backbone & 59.3 & 0.2 & 0.00 \\
+ MindJourney~\cite{mindjourney} & 66.7 & 2.5 & 12.34 \\
+ AVIC~\cite{avic} & 68.0 & 2.0 & 0.64 \\
+ AVIC-R~\cite{avic} & 69.3 & 4.8 & 3.03 \\
\bottomrule
\end{tabular}
\end{table}

SpatialUncertain v2~\cite{spatialuncertain} evaluates observational uncertainty with a different
label policy. Under complete occlusion, target-dependent questions require \emph{Cannot determine};
visibility questions can remain answerable. Under perspective conflict, a physical-3D-consistent
answer is correct and abstention is a separate category. An evaluation adapter must preserve those
classes. A spatial event probability cannot be scored by mapping \emph{Cannot determine} to a false
event. For this protocol, the adapter reserves an abstention action, chooses its threshold on
source-only calibration scenes, and reports answerability accuracy and conditional selective risk
separately. Benchmark versions and query sets must match when importing published results.

\subsection{Cost And Reuse}
\label{app:cost}
\begin{table}[t]
\centering\small\setlength{\tabcolsep}{3pt}
\caption{\textbf{Training and inference cost profile.} Eight H100 80\,GB accelerators, BF16, synchronized timing, and hypothesis microbatch 128. Teacher and student costs are separate; the codebook is shared. Cold totals include all displayed language-path components, while geometric execution is a separate readout. Entries give five-seed mean$_{\pm\mathrm{SD}}$.}
\label{tab:runtime}
\begin{tabular}{@{}lrl@{}}
\toprule
Component & Time & Unit \\
\midrule
\multicolumn{3}{l}{\textit{Training: total accelerator hours}} \\
Shared Shape Codebook & $96.0_{\scriptscriptstyle\pm 1.6}$ & GPU h \\
Teacher, Both Stages & $1096.0_{\scriptscriptstyle\pm 25.3}$ & GPU h \\
Scene NLL Student & $784.0_{\scriptscriptstyle\pm 15.8}$ & GPU h \\
Scene NLL, 40 Epochs & $1568.0_{\scriptscriptstyle\pm 31.6}$ & GPU h \\
Answer-Distillation Student & $540.0_{\scriptscriptstyle\pm 12.6}$ & GPU h \\
RevealBelief Student & $1864.0_{\scriptscriptstyle\pm 34.8}$ & GPU h \\
\midrule
\multicolumn{3}{l}{\textit{Inference: one observation set, eight questions}} \\
Encode Acquired Images & $0.080_{\scriptscriptstyle\pm 0.003}$ & s \\
Sample 1,024 States & $43.60_{\scriptscriptstyle\pm 0.71}$ & s \\
Execute Eight Questions & $0.026_{\scriptscriptstyle\pm 0.002}$ & s \\
Render 128 Views & $0.460_{\scriptscriptstyle\pm 0.016}$ & s \\
Soft Language Readout & $11.80_{\scriptscriptstyle\pm 0.32}$ & s \\
Cold Language Total & $55.94_{\scriptscriptstyle\pm 1.05}$ & s \\
Cached Geometric Question & $0.0034_{\scriptscriptstyle\pm 0.0002}$ & s \\
Cached Language Question & $1.46_{\scriptscriptstyle\pm 0.05}$ & s \\
\bottomrule
\end{tabular}
\end{table}

The probability bank and the render subset incur different costs. Sampling 1,024 autoregressive
scenes is charged to the initial observation set, although executing geometric predicates on those
states is inexpensive. The 32-root language path renders 128 images once. An eight-question batch
shares those images and one solver call per root, giving 16 renders per question. Cached geometric
questions need neither new roots nor renders. A new acquisition invalidates both banks and their
caches.

The runtime profile uses eight H100 80\,GB accelerators, BF16, hypothesis microbatches of 128, and
synchronized timing after 20 warm-up batches. Initial-query timing includes encoding, sampling,
rendering when used, and solver work. Cached-query timing retains the scene and image caches; it
includes new answer decoding. Training records separate the shared surface codebook, teacher
fitting, and student fitting. The 40-epoch Scene NLL control doubles student updates and uses 1,568
GPU-hours. Our 1,096 teacher hours and 1,864 student hours total 2,960 GPU-hours, excluding the
shared codebook. This is an update-count control; target sampling and the joint estimator contribute
to its different accelerator cost. These costs make explicit which work is amortized across
questions and which work returns after a new acquisition.

\section{Finite-World Diagnostics and Failure Analysis}
\label{app:diagnostics}
\begin{figure*}[t]
\centering
\includegraphics[width=\textwidth]{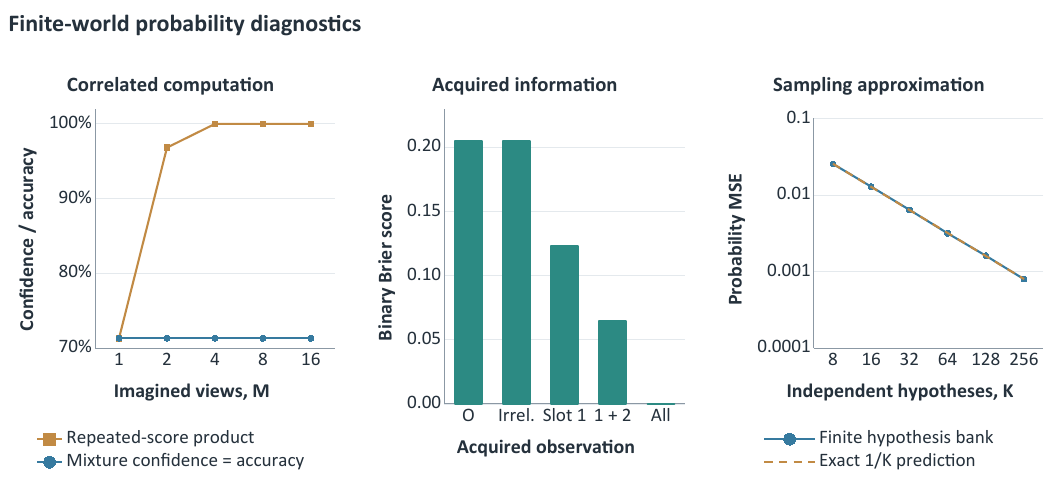}
\caption{\textbf{Exact finite-world mechanisms.} A specified 24-world distribution separates
  repeated-score sharpening, acquired information, and scene-sampling error. These calculations have
  no visual-perception error; each panel varies the stated probabilistic operation while keeping the
  others fixed.}
\label{fig:exact-diagnostics}
\end{figure*}

\subsection{A Completely Specified Scene Distribution}
The finite reference space contains all 24 permutations of four object identities over four slots.
The slot coordinates $(x,d)$ are $(-1.6,0.7)$, $(-0.5,2.6)$, $(0.7,1.3)$, and $(1.9,3.8)$. We
evaluate six left--right relations, six front--back relations, and twelve relative-distance
comparisons. Every predicate is a deterministic function of scene geometry. The prior places mass
$0.45$ on the identity permutation and $0.55/23$ on each other permutation. This prior differs from
the balanced four-alternative RevealRooms control.

An opaque initial mask hides all identity-bearing regions. Its RGB array is identical across the 24
worlds. The irrelevant view changes a common background but preserves the same observation
equivalence class. Revealing one slot, two slots, and the entire configuration yields 4, 12, and 24
observation classes. The likelihood is the exact deterministic observation map, and the posterior is
obtained by summing prior mass over compatible worlds. No neural perception is involved in these
calculations. The schematic renderer checks equality of the actual image arrays rather than assuming
it from symbolic labels.

\begin{table}[t]
\centering\small
\caption{\textbf{Actual-view updates in the finite reference model.} Accuracy is averaged over the 24 binary queries; entropy describes the latent scene distribution in nats. These are exact expectations over observation outcomes.}
\label{tab:finiteviews}
\begin{tabular}{@{}lrrr@{}}
\toprule
Observation & Acc. (\%) & Brier & Entropy\\
\midrule
Initial occluded & 71.30 & 0.2046 & 2.4127\\
Irrelevant real view & 71.30 & 0.2046 & 2.4127\\
Reveal one slot & 80.47 & 0.1230 & 1.2563\\
Reveal two slots & 88.04 & 0.0643 & 0.4594\\
Full reveal & 100.00 & 0.0000 & 0.0000\\
\bottomrule
\end{tabular}
\end{table}

\subsection{Repeated Scores And Correlated Samples}
The deliberately misspecified product rule takes a nonuniform scene mass vector $\pi$ and replaces
it by $\pi_k^M/\sum_\ell\pi_\ell^M$. This is not a model of a particular published system. It
isolates what happens when computational agreement is repeatedly interpreted as an independent
likelihood. With $M=4$, scene-set coverage at nominal 90\% falls from $90.43\%$ to $45.00\%$, and
the selected set shrinks from twenty worlds to one. Query point accuracy is unchanged. For a uniform
mass vector, exponentiation does not change the distribution, providing a negative control.

A second calculation samples $K$ independent roots and renders each one $M$ times. The empirical
answer probability is unchanged when every root is duplicated equally. However, treating all $KM$
copies as independent narrows an interval for the Monte Carlo estimate incorrectly. We use a Wilson
interval for one binary query and 10,000 independent particle-bank repetitions per $K$, with random
seed 20261108. These repetitions quantify numerical approximation, not variation across fitted
neural models.

\begin{figure*}[t]
\centering
\includegraphics[width=\textwidth]{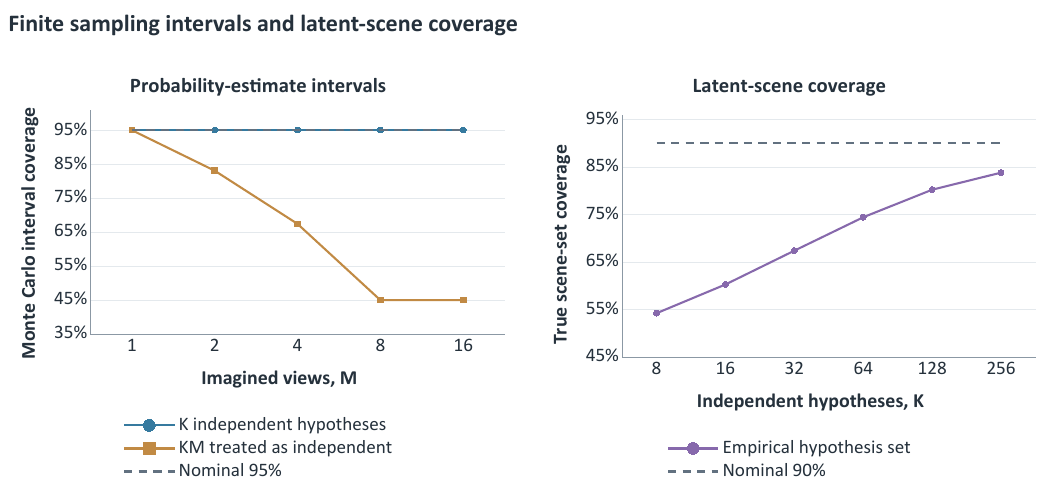}
\caption{\textbf{Two different notions of coverage.} \textbf{Left:} 95\% confidence intervals for
  the Monte Carlo estimate of an answer probability, with $K=32$. Renders of one sampled root do not
  increase the number of independent scene samples. \textbf{Right:} actual coverage of an empirical
  90\% credible set for the latent scene. Even correctly grouped roots can omit important states, so
  duplication invariance does not ensure predictive coverage.}
\label{fig:coverage}
\end{figure*}

\begin{table}[t]
\centering\small
\caption{\textbf{Monte Carlo interval coverage.} $K=32$ independent roots; percentages across 10,000 repetitions. Both rules use the same empirical probability estimate. Only the assumed sample count changes.}
\label{tab:mccoverage}
\begin{tabular}{@{}rrrr@{}}
\toprule
$M$ & Correct count $K$ & False count $KM$ & Naive width\\
\midrule
1 & 95.07 & 95.07 & 0.295\\
2 & 95.07 & 83.13 & 0.213\\
4 & 95.07 & 67.47 & 0.152\\
8 & 95.07 & 45.01 & 0.108\\
16 & 95.07 & 45.01 & 0.077\\
\bottomrule
\end{tabular}
\end{table}

A credible set for the actual world answers a different question. We sort the empirical root mass
and include the smallest prefix reaching the specified level, with stable tie breaking. At $K=32$,
the empirical 90\% set has true coverage $67.43\%$; at $K=256$, it has $83.85\%$. A finite bank
therefore requires its own coverage analysis. Renaming it a posterior does not remove sampling
error. The exact enumerated set covers $90.43\%$ because its discrete mass crosses the nominal
threshold.

\begin{table}[t]
\centering\small
\caption{\textbf{Root budget in the finite reference.} Each entry averages 10,000 banks, with $M=8$ repeated renders per root. Intervals target 95\% coverage of the Monte Carlo event probability; scene sets target 90\% latent mass.}
\label{tab:root_budget}
\begin{tabular}{@{}rrrrr@{}}
\toprule
& \multicolumn{2}{c}{Interval coverage (\%)} & \multicolumn{2}{c}{Scene set}\\
\cmidrule(lr){2-3}\cmidrule(l){4-5}
$K$ & Grouped & Naive & Cov. (\%) & Size\\
\midrule
8 & 95.32 & 54.90 & 54.25 & 5.03 \\
32 & 95.07 & 45.01 & 67.43 & 10.38 \\
128 & 94.91 & 49.85 & 80.26 & 15.75 \\
256 & 94.66 & 51.31 & 83.85 & 17.25 \\
\bottomrule
\end{tabular}
\end{table}

\subsection{Unknown Prior Shift}
We change the test distribution while preserving the observation map. Under a uniform test prior, a
system retaining the spiked training prior has query Brier score $0.2954$, whereas the informed
posterior attains $0.2500$. Moving the $0.45$ mode to a different permutation gives $0.3256$ versus
$0.2046$. An initial image is identical under these priors, so the shift cannot be inferred from
that image. The informed comparator receives the new prior explicitly and is not a deployable
adaptation method.

\begin{table}[t]
\centering\small
\caption{\textbf{Prior misspecification.} Query Brier score for an unchanged observation model. ``Informed'' is given the test prior; ``frozen'' retains the training prior.}
\label{tab:priorshift}
\begin{tabular}{@{}llrr@{}}
\toprule
Test prior & Observation & Frozen & Informed\\
\midrule
Uniform & Initial & 0.2954 & 0.2500\\
Uniform & One slot & 0.2058 & 0.1806\\
Uniform & Two slots & 0.1112 & 0.1042\\
Moved mode & Initial & 0.3256 & 0.2046\\
Moved mode & One slot & 0.1951 & 0.1230\\
Moved mode & Two slots & 0.1082 & 0.0643\\
Either & Full reveal & 0.0000 & 0.0000\\
\bottomrule
\end{tabular}
\end{table}

Revealing observations reduce prior dependence when they distinguish the candidate worlds. Complete
identification eliminates the prior difference in this deterministic example. This does not imply
that any real view eliminates prior bias: an irrelevant image has a constant likelihood across
alternatives and cannot help.

\subsection{Failure Taxonomy}
\paragraph{Missing modes.}
A bank containing only world A cannot assign probability to world B, even if every repeated render
is grouped correctly. Posterior refresh helps only when the scene model can propose B. The system
does not include an unvalidated ``other-world mass'' that purports to quantify all missing
realities.

\paragraph{Teacher bias.}
A revealing teacher may infer the wrong object identity or collapse unresolved alternatives. The
partial-view student can inherit that error. The teacher bound in \cref{app:theory} makes this
dependence explicit. Calibration on the teacher's own predictions cannot establish accuracy relative
to the actual scene.

\paragraph{Camera and representation error.}
Camera errors and geometry quantization can reverse spatial relations despite correct scene weights.
The predicate margins are fixed before testing; changing them afterward would change the task.
Out-of-catalog objects require separate transfer analysis.

\paragraph{Reasoner error.}
A VLM can misread correct geometry. The analytic executor and the direct scene-code solver provide
references for assessing whether rendered views help the approximate branch computation.

\paragraph{Decision inconsistency.}
Coherent marginals can yield cyclic thresholded answers. Requiring a realizable signature can reduce
marginal accuracy (\cref{app:theory}); probability quality and decision feasibility remain separate
criteria.

\section{Worked Probability Examples}
\label{app:worked}
The following finite distributions illustrate why marginal calibration, joint probability quality,
and conditional solving require different tests. Each row is defined by a probability vector or a
channel parameter. The examples complement the learned-model comparisons without assigning method
names to prescribed distributions.

\subsection{Nonuniform Conditional Probabilities}
\label{app:nonuniform}
Consider three queries: red left of blue, green in front of blue, and blue right of red. Their four
realizable signatures are:
\begin{equation}
 F=\begin{bmatrix}0&0&0\\0&1&0\\1&0&1\\1&1&1\end{bmatrix}.
 \label{eq:app-four-signatures}
\end{equation}
The first and third entries express the same semantic fact. The middle fact can vary independently
across geometry, although their population frequencies need not factor. Define three equally
weighted probability contexts with priors
\begin{align*}
 \pi_1&=(.55,.10,.10,.25),\\
 \pi_2&=(.55,.25,.15,.05),\\
 \pi_3&=(.05,.15,.25,.55).
\end{align*}
Their event marginals are $(.35,.35,.35)$, $(.20,.30,.20)$, and $(.80,.70,.80)$. These probability
contexts are independent of the Natural scene protocol.

Let $u=(1/4,1/4,1/4,1/4)$ and define $q_{c,\alpha}=(1-\alpha)\pi_c+\alpha u$.
\cref{tab:nonuniform-calibration} varies only $\alpha$. Conditional probability error (CPE) is the
mean absolute event-probability error over context--query pairs. The resulting scores isolate prior
information relative to a constant-$0.5$ forecast.

\begin{table}[t]
\centering\small
\caption{\textbf{Shrinkage toward a uniform distribution.} The parameter $\alpha$ fully defines each row. CPE is in percentage points.}
\label{tab:nonuniform-calibration}
\begin{tabular}{@{}lrrr@{}}
\toprule
$\alpha$ & Binary Brier & Binary NLL & CPE\\
\midrule
1.00 & .2500 & .6931 & 22.78\\
0.28 & .1980 & .5850 & 6.38\\
0.14 & .1947 & .5769 & 3.19\\
0.06 & .1938 & .5745 & 1.37\\
0.00 & .1936 & .5740 & 0.00\\
\bottomrule
\end{tabular}
\end{table}

\subsection{Equal Marginals, Different Dependencies}
Fix $\pi=\pi_1$ from the previous example. The family:
\begin{equation}
 q(a)=(a,.65-a,.65-a,a-.30),\quad .30\le a\le .65
\end{equation}
has exactly the same marginals $(.35,.35,.35)$ for every valid $a$. The two distinct facts are
independent at $a=.4225$. All vectors preserve the first/third logical equivalence. Their binary
Brier is $.2275$ and their invalid-signature mass is zero, yet their joint scores differ.

\begin{table}[t]
\centering\small
\caption{\textbf{A dependence example with exact marginals.} Rows are specified by $a$, not by model names. The true distribution has $a=.55$.}
\label{tab:joint-signature-probabilities}
\begin{tabular}{@{}lrr@{}}
\toprule
$a$ & Joint Brier & Joint NLL\\
\midrule
.4225 (independence) & .6800 & 1.2949\\
.49 & .6294 & 1.1740\\
.52 & .6186 & 1.1462\\
.53 & .6166 & 1.1407\\
.55 (true) & .6150 & 1.1359\\
\bottomrule
\end{tabular}
\end{table}

Joint Brier equals $1-\|\pi\|_2^2+\|q(a)-\pi\|_2^2$. The first term is irreducible outcome
uncertainty; the second measures the distribution error that marginal scores miss.

\subsection{A Prescribed Readout Channel}
Hold scene mass fixed at $q(.53)=(.53,.12,.12,.23)$. Within each of four hypotheses, a channel flips
each of the two distinct facts independently with probability $\epsilon$, then assigns confidence
$c$ to its hard decision. The third query copies the first. The channel's joint kernel is the
product over the two distinct facts, followed by this copy. Averaging these kernels with $q(.53)$
preserves the declared equivalence.

We enumerate all $2^8$ flip patterns, reuse the prediction bank across observationally equivalent
outcomes, and average their losses under the true prior. Scoring the averaged forecast would omit
channel variance.

\begin{table}[t]
\centering\small
\caption{\textbf{Readout-channel parameters.} The pair $(\epsilon,c)$ defines each stochastic channel. Root accuracy concerns a candidate's facts; answer accuracy concerns the hidden actual world. These are exact channel expectations.}
\label{tab:branch-readout}
\begin{tabular}{@{}lrrr@{}}
\toprule
$(\epsilon,c)$ & Root acc. & Answer acc. & Brier\\
\midrule
$(0,1)$ & 100 & 65.0 & .2275\\
$(.12,.90)$ & 88 & 61.4 & .2555\\
$(.06,.94)$ & 94 & 63.2 & .2445\\
$(.14,.90)$ & 86 & 60.8 & .2595\\
\bottomrule
\end{tabular}
\end{table}

Lower channel error improves the answer at fixed scene mass; the deterministic executor supplies the
exact reference. These channel parameters are stipulated independently of the learned-solver
experiment.

\subsection{Teacher Contamination}
Retain $\pi=(.55,.10,.10,.25)$ and define $r=(.05,.15,.25,.55)$. A contaminated mean teacher target
is $T_\epsilon=(1-\epsilon)\pi+\epsilon r$. An unrestricted distilled student equals this target, so
its expected joint Brier under the original population is:
\begin{equation}
 1-\|\pi\|_2^2+\epsilon^2\|r-\pi\|_2^2
 =.615+.365\epsilon^2.
\end{equation}
At $\epsilon=0,.1,.25,.5$, joint Brier is $.6150,.6187,.6378,.7063$. The corresponding binary Brier
values are $.2275,.2293,.2385,.2715$. The population does not change. This isolates teacher bias,
whereas \cref{app:diagnostics} changes the true prior while retaining the predictor.

\end{document}